\documentclass[10pt,twocolumn,letterpaper]{article}
\PassOptionsToPackage{numbers, sort, compress}{natbib}
\usepackage{cvpr}   
\usepackage[breaklinks=true,bookmarks=false,colorlinks,citecolor=green]{hyperref}
\usepackage{comment}
\usepackage{amsmath}

\usepackage[capitalize,noabbrev]{cleveref}

\usepackage[utf8]{inputenc} 
\usepackage[T1]{fontenc}    
\usepackage{url}            
\usepackage{booktabs}       
\usepackage{amsfonts}       
\usepackage{nicefrac}       
\usepackage{microtype}      
\usepackage{xcolor}         

\usepackage{epsfig}
\usepackage{tabularx}
\usepackage{graphicx}
\usepackage{amssymb}
\usepackage{bbm}
\usepackage{color}
\usepackage{wrapfig}
\usepackage{subcaption}
\usepackage{float}
\usepackage{makecell}
\usepackage[numbers,sort,compress]{natbib}
\usepackage{adjustbox}

\usepackage{multirow}
\usepackage{inconsolata}
\usepackage{comment}
\usepackage{xspace}

\newcommand{\ourtasklong}{openable part detection\xspace}
\newcommand{\ourtask}{OPD\xspace}
\newcommand{\ourdataset}{\ourtask dataset\xspace}
\newcommand{\ourdatareal}{OPDReal\xspace}
\newcommand{\ourdatacad}{OPDSynth\xspace}
\newcommand{\pmdataset}{PartNet-Mobility\xspace}
\newcommand{\NUMOBJECTS}{683\xspace}
\newcommand{\NUMOBJCATS}{11\xspace}

\newcommand{\NUMPARTS}{1343\xspace}

\newcommand{\method}{\textsc{OpdRcnn}\xspace}
\newcommand{\opdnetoc}{\textsc{OpdRcnn-O}\xspace}
\newcommand{\opdnetcc}{\textsc{OpdRcnn-C}\xspace}
\newcommand{\opdnetocs}{\textsc{OpdRcnn-O}\xspace}

\newcommand{\ancsh}{\textsc{ANCSH}\xspace}
\newcommand{\pnetopd}{\textsc{OpdPN}\xspace}
\newcommand{\randmot}{\textsc{RandMot}\xspace}
\newcommand{\mostfreq}{\textsc{MostFreq}\xspace}

\newcommand{\gtboxpart}{\textsc{GT Box2DPart}\xspace}
\newcommand{\gtpose}{\textsc{GT Pose}\xspace}
\newcommand{\gtboxpartpose}{\textsc{GT Box2DPartPose}\xspace}

\newcommand{\drawer}{\texttt{drawer}\xspace}
\newcommand{\door}{\texttt{door}\xspace}
\newcommand{\lidd}{\texttt{lid}\xspace}
\newcommand{\mttrans}{\texttt{prismatic}\xspace}
\newcommand{\mtrot}{\texttt{revolute}\xspace}

\newcommand{\axis}{\textbf{A}\xspace}
\newcommand{\orig}{\textbf{O}\xspace}
\newcommand{\mtype}{\textbf{M}\xspace}
\newcommand{\partdet}{\textbf{PDet}\xspace}
\newcommand{\motiondet}{\textbf{MDet}\xspace}

\newcommand{\ap}{\ensuremath{\text{AP}}}
\newcommand{\apbb}{\ensuremath{\text{AP}^\text{bb}}}

\newcommand{\R}{\ensuremath{\mathbb{R}}}

\newcommand{\indicator}{\ensuremath{\mathbbm{1}}}

\newcommand{\denselist}{\itemsep 0pt\parsep=0pt\partopsep 0pt\vspace{-\topsep}}
\newcommand{\mypara}[1]{\vspace{2pt}\noindent\textbf{#1}}

\newcommand{\losshuber}{\ensuremath{L_\text{smoothL1}}}
\newcommand{\lossce}{\ensuremath{L_\text{CE}}}
\newcommand{\lossbce}{\ensuremath{L_\text{BCE}}}

\newcommand\markbest[1]{#1}

\newcolumntype{Y}{>{\centering\arraybackslash}X}
\newcommand\imgclip[2]{\adjincludegraphics[Clip={#1\width} {#1\height} {#1\width} {#1\height}]{#2}}

\begin{document}
\title{\ourtask: Single-view 3D Openable Part Detection}

\author{
Hanxiao Jiang, Yongsen Mao, Manolis Savva, Angel X. Chang\\
Simon Fraser University \\
\href{https://3dlg-hcvc.github.io/OPD/}{3dlg-hcvc.github.io/OPD/}
}

\pagestyle{plain}

\twocolumn[{
\maketitle
\begin{center}
\includegraphics[width=\linewidth]{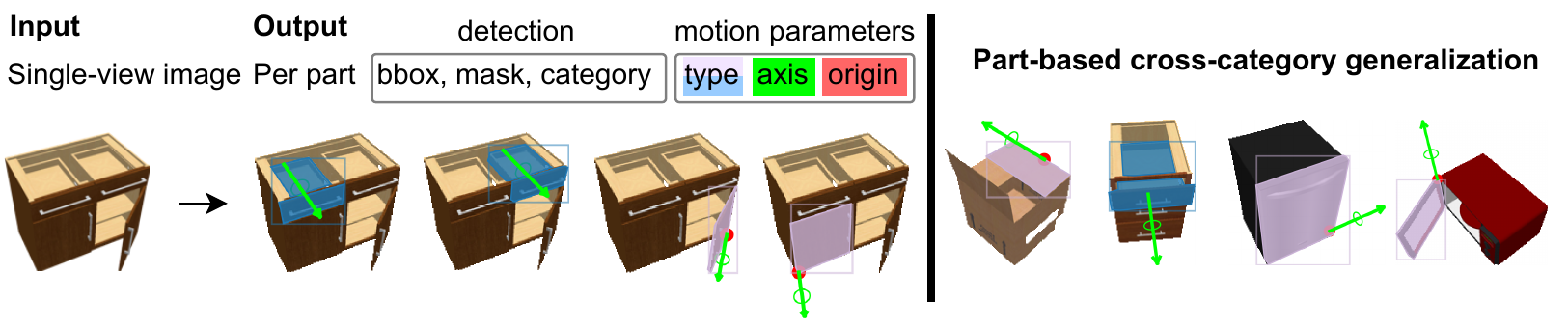}
\captionof{figure}{
We address the task of \ourtasklong (\ourtask).
The input is a single view image, and the outputs are detected openable parts as well as their motion parameters.
We design \method: a neural architecture for this task that operates at the part level, allowing generalization across diverse object categories.
}
\label{fig:inputoutput}
\end{center}
}]

\begin{abstract}
We address the task of predicting what parts of an object can open and how they move when they do so. The input is a single image of an object, and as output we detect what parts of the object can open, and the motion parameters describing the articulation of each openable part. To tackle this task, we create two datasets of 3D objects: \ourdatacad based on existing synthetic objects, and \ourdatareal based on RGBD reconstructions of real objects. We then design \method, a neural architecture that detects openable parts and predicts their motion parameters. Our experiments show that this is a challenging task especially when considering generalization across object categories, and the limited amount of information in a single image. Our architecture outperforms baselines and prior work especially for RGB image inputs.

\end{abstract}
\section{Introduction}
There is increasing interest in training embodied AI agents that interact with the world based on visual perception.
Recently, \citet{batra2020rearrangement} introduced rearrangement as a challenge bringing together machine learning, vision, and robotics.
Common household tasks such as ``put dishes in the cupboard'' or ``get cup from the cabinet'' can be viewed as object rearrangement.
A key challenge in such tasks is identifying which parts can be opened and how they open.

To address this problem, we introduce the \textit{\ourtasklong} (\ourtask) task, where the goal is to identify openable parts, and their motion type and parameters (see \Cref{fig:inputoutput}).
We focus on predicting the openable parts (``door'', ``drawer'', and ``lid'') of an object, and their motion parameters from a single RGB image input. 
More specifically, we focus on container objects (e.g. cabinets, ovens, etc).
Containers need to be opened to look for hidden objects to reach, or to place away objects.
Methods that can identify what can be opened and how offer a useful abstraction that can be used in more complex tasks~\cite{shridhar2020alfred,weihs2021visual,mittal2021articulated,szot2021habitat,srivastava2022behavior}.

\begin{table*}[t]
\caption{
Summary of prior work on motion parameter estimation (motion type, axis, and rotation origin).
We indicate the input modality, whether the method has cross-category generalization (CC),
whether part segmentations are predicted (Seg) as opposed to using ground-truth parts, whether object pose (OP) and part state (PS) are predicted.
Most prior work takes point cloud (PC) inputs.
In contrast, our input is single-view images (RGB, D, or RGB-D).
}
\centering
{
\begin{tabular}{@{}l @{\hspace{1em}} l @{\hspace{1em}} cccc @{\hspace{1em}} rrr@{}}
\toprule
Method & Input & CC & Seg & OP & PS & \#cat & \#obj & \#part\\
\midrule
Snapshot~\cite{hu2017learning} & 3D mesh & & & &  & & 368 & 368\\
Shape2Motion~\cite{wang2019shape2motion} & PC & \checkmark & \checkmark& & & 45 & 2440 & 6762\\
RPMNet~\cite{yan2019rpm} & PC & \checkmark & \checkmark & & & & 969 & 1420 \\
DeepPartInduction~\cite{yi2019deep} & Pair of PCs & & \checkmark& & & 16 & 16881 & \\
MultiBodySync~\cite{huang2021multibodysync} & Multiple PCs & & \checkmark& & & 16 & & \\
ScrewNet~\cite{jain2020screwnet} & Depth video & \checkmark & & & & 9 & 4496 & 4496\\
\citet{liu2020nothing} & RGB video & \checkmark & \checkmark & & & 3 & & \\
ANCSH~\cite{li2020category} & Single-view PC &  & \checkmark& \checkmark & \checkmark & 5 & 237 & 343\\
\citet{abbatematteo2019learning} & RGB-D & & \checkmark & &  & 5 & 6 & 8 \\
VIAOP~\cite{zeng2020visual}  & RGB-D & \checkmark & & & & 6 & 149 & 618\\
\midrule
\method (ours) & Single-view RGB(-D) & \checkmark & \checkmark& \checkmark & \checkmark & \NUMOBJCATS & \NUMOBJECTS & \NUMPARTS\\
\bottomrule
\end{tabular}
}

\label{tab:related}
\end{table*}

Prior work~\cite{hu2018functionality,yan2019rpm,li2020category} has looked at predicting motion types and parameters from 3D point clouds.
Point-cloud based methods rely on depth information, often sampled from a reconstructed mesh that aggregates information from multiple views.
In addition, \citet{li2020category} assumes that the kinematic chain and the part structure of the objects are given.
To handle different kinematic structures, they train a separate model for each structure.
This approach is not scalable to the wide variety of structures found in the real world, even for objects in the same category.
For example, cabinets may have two, three or four drawers.

We study the task of identifying openable parts and motion parameters from a single-view image, and investigate whether a structure-agnostic image-based approach can be effective at this task.
Our approach \method extends instance segmentation networks (specifically MaskRCNN) to identify openable parts and predict their motion parameters.
This structure-agnostic approach allows us to more easily achieve cross-category generality.
That is, a single model can tackle instances from a variety of object categories (e.g. cabinets, washing machines).
In addition, we investigate whether depth information is helpful for this task.

In summary, we make the following contributions:
i) we propose the \ourtask task for predicting openable parts and their motion parameters from single-view images;
ii) we contribute two datasets of objects annotated with their openable parts and motion parameters: \ourdatacad and \ourdatareal; and
iii) we design \method: a simple image-based, structure-agnostic neural architecture for openable part detection and motion estimation across diverse object categories.
We evaluate our approach with different input modalities and compare against baselines to show which aspects of the task are challenging.
We show that depth is not necessary for accurate part detection and that we can predict motion parameters from RGB images.
Our approach significantly outperforms baselines especially when requiring both accurate part detection and part motion parameter estimation.

\begin{figure*}[ht]
\centering
\setkeys{Gin}{width=\linewidth}
\begin{tabularx}{\textwidth}{Y@{\hspace{1mm}}
Y@{\hspace{1mm}} | Y@{\hspace{1mm}} Y@{\hspace{1mm}} Y}
\includegraphics{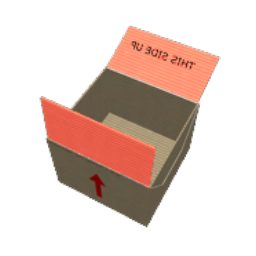} &
\includegraphics{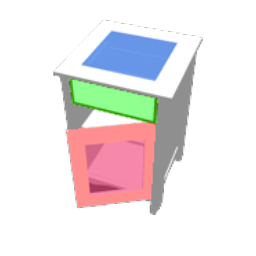} &
\includegraphics{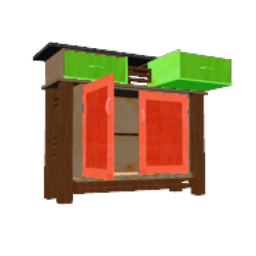} &
\includegraphics{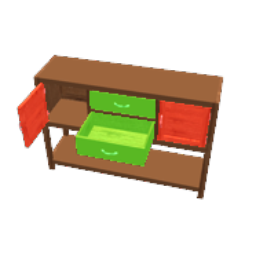} &
\includegraphics{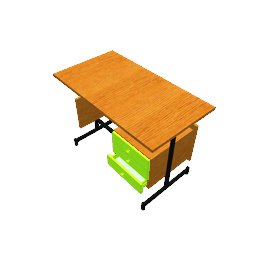} \\
\includegraphics{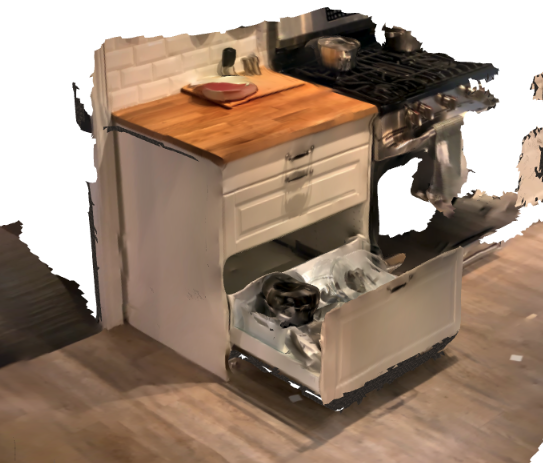} &
\includegraphics{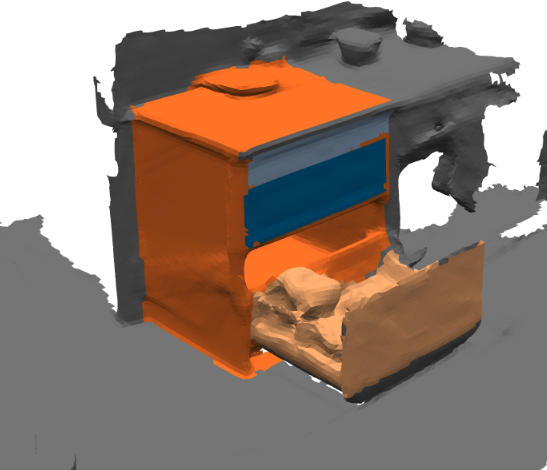} &
\includegraphics{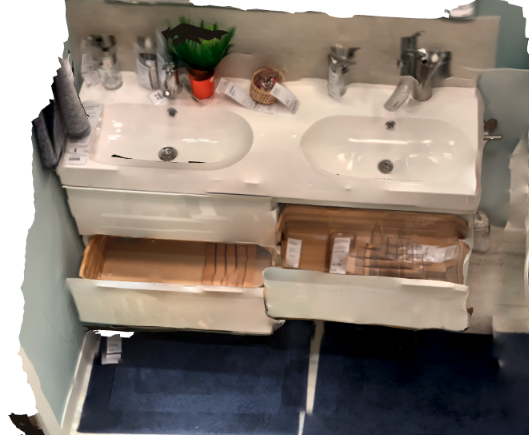} &
\includegraphics{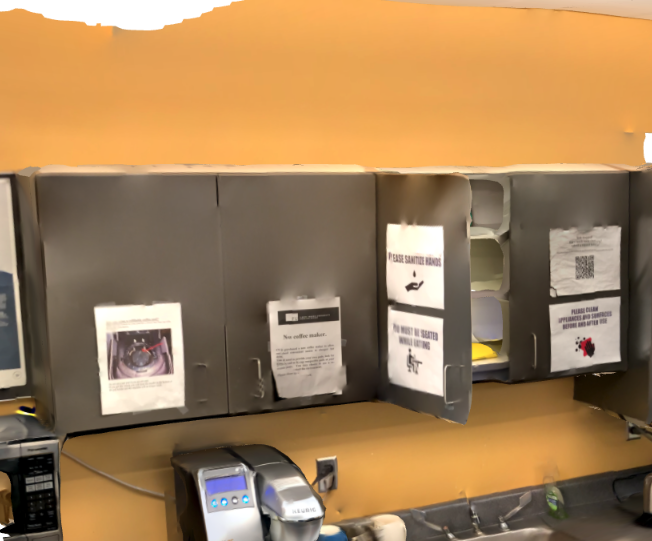} &
\includegraphics{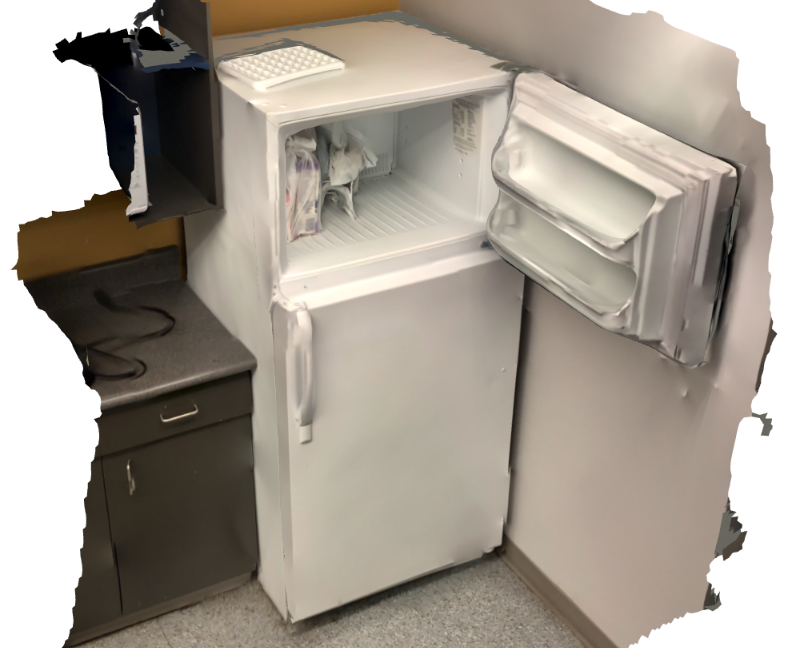} 
\end{tabularx}
\caption{Example articulated objects from the \ourdatacad dataset and the \ourdatareal dataset. The first row is from \ourdatacad. Left: different openable part categories (\lidd, in orange, \drawer in green, \door in red). Right: Cabinet objects with different kinematic structures and varying numbers of openable parts. The second row is from our \ourdatareal dataset. Left: reconstructed cabinet and its semantic part segmentation. Right: example reconstructed objects from different categories.
}
\label{fig:data-examples}
\end{figure*}

\section{Related Work}

\mypara{Part segmentation and analysis.}
Part segmentation has been widely studied in the vision and graphics communities for both 2D images and 3D representations.
Much of the prior work relies on part annotations in 2D images~\cite{chen2014detect,zhou2017scene}, or 3D CAD models with semantically labeled part segmentations~\cite{yi2016scalable,wang2018learning,mo2019partnet, wang2019shape2motion, yan2019rpm}.
Annotated 3D part datasets have been used to study part segmentation by rendering images~\cite{varol2017learning} or directly in 3D (e.g. meshes or point clouds)~\cite{yi2016scalable,mo2019partnet}.
These datasets have fostered the development of different methods to address part segmentation on images~\cite{wang2015semantic,wang2015joint,xia2017joint}.
Many of these focus on human/animal part segmentation, and use hierarchical methods to parse both objects and the parts.
Beyond part segmentation, \cite{lu2018beyond} further enhanced the PASCAL VOC Part dataset~\cite{chen2014detect} with state information.
In contrast, we take a simple approach and directly detect and segment the parts of interest using standard object instance segmentation methods~\cite{he2017mask, bochkovskiy2020yolov4}.

Unlike prior work, we focus on a small set of openable part categories.
Part datasets differ in what parts they focus on (e.g., human body parts, or fine vs coarse-grained object parts).
Determining the set of parts of interest can be tricky, as
the set of possible parts can be large and ill-defined, with what constitutes a part varying across object categories.
Therefore, we focus on a small set of openable parts in common household objects.
We believe this set of parts is a small, practical set that is important for object interaction.

\mypara{Articulated object motion prediction.}
Part mobility analysis is a long-standing problem in 3D computer graphics.
Early work~\cite{mitra2010illustrating} has focused on learning the part mobility in mechanical assemblies.
\citet{xu2009joint} used a mobility tree formalism to further explore object and part mobility in indoor 3D scenes.
More recent work\cite{sharf2014mobility} proposed a joint-aware deformation framework based on shape analysis and optimization to predict motion joint parameters.
Part mobility analysis has also been performed on sequences of RGBD scans~\cite{li2016mobility}.
More recently, there has been increasing interest in data-driven methods for studying articulated objects and estimating motion parameters\cite{wang2019shape2motion,li2020category}.
To support these data-driven approaches, there has been concurrent development of datasets of annotated part articulations for synthetic~\cite{yan2019rpm,wang2019shape2motion,xiang2020sapien} and reconstructed~\cite{martin2019rbo,liu2022akb} 3D objects.

\Cref{tab:related} summarizes the part mobility prediction tasks defined by this and other recent work.
\citet{hu2017learning} predict joint parameters given pre-segmented 3D objects.
Other work predicts segmentation together with motion parameters, for 2.5D inputs~\cite{yan2019rpm, yi2019deep}, 3D point clouds~\cite{wang2019shape2motion} or for sequences of RGBD scans \cite{li2016mobility}.
\citet{abbatematteo2019learning,li2020category} have the most similar task setting with our work, predicting the part segmentation, part pose, and joint parameters from a single view image.  
However, both require depth as input and knowledge of the kinematic chain of each object.
They require training a separate model for each object category, where the object category is defined as having the same structure (i.e. same kinematic chain).
This means that different models need to be trained for cabinets with 3 drawers and cabinets with 4 drawers.
In contrast, we identify all openable parts of an object in an input RGB image, without assuming a specific structure with given number of parts.
This allows us to train a single model that generalizes across categories.
Note that \citet{abbatematteo2019learning} also use MaskRCNN for segmentation, but they do not analyze or report the part segmentation and detection performance of their model.

More recent work has started to explore training of single models for motion prediction across categories and structures~\cite{jain2020screwnet,jain2020screwnet,liu2020nothing,huang2021multibodysync}.
~\citet{zeng2020visual} proposed an optical flow-based approach on RGB-D images given segmentation masks of the moving part and fixed part.
They evaluate only on ground truth segmentation and do not investigate how part segmentation and detection influences the accuracy of motion prediction.
Others have proposed to predict articulated part pose from depth sequences~\cite{jain2020screwnet}, image video sequences~\cite{liu2020nothing}, or synchronizing multiple point clouds~\cite{huang2021multibodysync}.
In contrast, we focus on single-view image input and show that even without depth information, we can accurately predict motion parameters.

\begin{table*}[h]
\caption{\ourdataset statistics. We create two datasets of objects with openable parts: \ourdatacad and \ourdatareal. The datasets contain various object categories with potentially multiple openable parts. We annotate the semantic part segmentation and articulation parameters on 3D polygonal meshes, allowing us to generate many views of each object with ground truth.}
\resizebox{\linewidth}{!}{
\begin{tabular}{@{} ll rrrrrrrrrrr @{}}
\toprule
 & & \multicolumn{11}{c}{Category} \\
\cmidrule(l{0pt}r{0pt}){3-13}
             &         & Storage & Table & Bin & Fridge & Microwave & Washer & Dishwasher & Oven & Safe & Box & Suitcase \\
\multirow{3}{*}{\ourdatacad}  & Objects & 366 &  76 & 69 & 42 & 13 & 17 & 41 & 24 & 30 & 28 & 7 \\
             & Parts   & 809 & 187 & 75 & 68 & 13 & 17 & 42 & 36 & 30 & 59 & 7 \\
             & Images  & 89300 & 20600 & 9225 & 7850 & 1625 & 2125 & 5225 & 4200 & 3750 & 6600 & 875  \\
\cmidrule(l{0pt}r{0pt}){2-13}
\multirow{3}{*}{\ourdatareal} & Objects & 231 &  16 &  7 & 12 &  7 &  3 &  3 &  5 &  - &  - & - \\
             & Parts   & 787 & 35 & 7 & 27 & 7 & 3 & 3 & 6 & - & - & - \\
             & Images  & 27394 & 1175 & 474 & 1321 & 570 & 159 & 186 & 253 & - & - & -  \\
\bottomrule
\end{tabular}
}
\label{tab:dataset-summary-stats}
\end{table*}

\section{Problem Statement}

Our input is a single RGB image $I$ of a single articulated object and the output is the set of openable parts $P = \{p_1 \ldots p_k\}$ (i.e. drawers, doors and lids) with their motion parameters $\Phi = \{\phi_1 \ldots \phi_k\}$.
\Cref{fig:inputoutput} illustrates the input and output for our task.
For each part $p_i = \{b_i, m_i, l_i\}$, we predict the 2D bounding box $b_i$, the segmentation mask $m_i$, and the semantic label $l_i \in \{\drawer, \door, \lidd\}$.
The motion parameters $\phi_i$ of each openable part specify the motion type $c_i \in \{\mttrans, \mtrot\}$, motion axis direction $a_i \in R^3$ and motion origin $o_i \in R^3$.
Specifically, we have $\phi_i=[c_i, a_i, o_i]$ for revolute joints (e.g., door rotating around a hinge), and $\phi_i=[c_i, a_i]$ for prismatic joints (e.g., drawer sliding out).
For simplicity, we assume that each part has only one motion.
\section{OPD Dataset}

To study our task, we collate two datasets of objects with openable parts, a dataset of synthetic 3D models (\ourdatacad) and a dataset of real objects scanned with RGB-D sensors (\ourdatareal).
For \ourdatacad, we select objects with openable parts from an existing dataset of articulated 3D models \pmdataset~\cite{xiang2020sapien}.
For \ourdatareal, we reconstruct 3D polygonal meshes for articulated objects in real indoor environments and annotate their parts and articulation information.
\Cref{tab:dataset-summary-stats} shows summary statistics of the number of objects and openable parts for each object category (see supplement for more detailed statistics).

\mypara{\ourdatacad.}
The \pmdataset dataset contains 14K movable parts across 2,346 3D objects from 46 common indoor object categories \cite{xiang2020sapien}.
We canonicalize the part names to `drawer', `door', or `lid'.
We select object categories with openable parts that can serve as containers.
For each category, we identify the openable parts and label them as \drawer, \door, or \lidd (see supplement).
Overall, we collated \NUMOBJECTS objects with \NUMPARTS parts over \NUMOBJCATS categories (see \Cref{tab:dataset-summary-stats}).
We then render several views of each object to produce RGB, depth, and semantic part mask frames.
Specifically, for each articulated object we render different motions for each part.
In one motion state all parts are set to the minimum value in their motion range.
Then for each part, we pick 3 random motion states and the maximum of the motion range, while the other parts remain at the minimum value.
We render five images for each motion state from different camera viewpoints, and each image is composited on four different randomly selected background images (see supplement for rendering details).

\mypara{\ourdatareal.}
To construct a dataset of real objects, we take 863 RGB-D video scans of indoor environments with articulated objects (residences, campus lounges, furniture showrooms) using iPad Pro 2021 devices.
We obtain polygonal mesh reconstructions from these scans using the Open3D~\cite{zhou2018open3d} implementation of RGB-D integration and \citet{waechter2014texturing}'s implementation of texturing.
We follow an annotate-validate strategy to filter and annotate this scanned data.
Specifically, we: 1) Annotate the model quality and filter the scans with bad quality (insufficient geometry to annotate articulations); 2) Annotate the semantic part segmentation; 3) Validate the semantic segmentation; 4) Annotate the articulation parameters for articulated parts; 5) Validate the articulations through animating the moving parts; 6) Annotate a `semantic OBB' (center at origin, semantic `up' and `front' axis direction) for each object; 7) Calculate consistent object pose (i.e. transformation between camera coordinates and object coordinates) based on semantic OBB.
After filtering and annotation, we have a total of 763 polygonal meshes for 284 different objects across 8 object categories.
\Cref{tab:dataset-summary-stats} shows the distribution over different object categories.

We then project the 3D annotations on the polygonal mesh back to the original RGB and depth frames from the scan videos.
We select around 100 frames from each video and project the segmentation mask in 3D back to 2D.
The articulation parameters in world coordinate are also projected to the camera coordinates of each frame.
When selecting frames, we sample one frame every second ensuring that at least 1\% of pixels belong to an openable part and at least 20\% of parts are visible.

\begin{figure*}[t]
\includegraphics[width=1\linewidth]{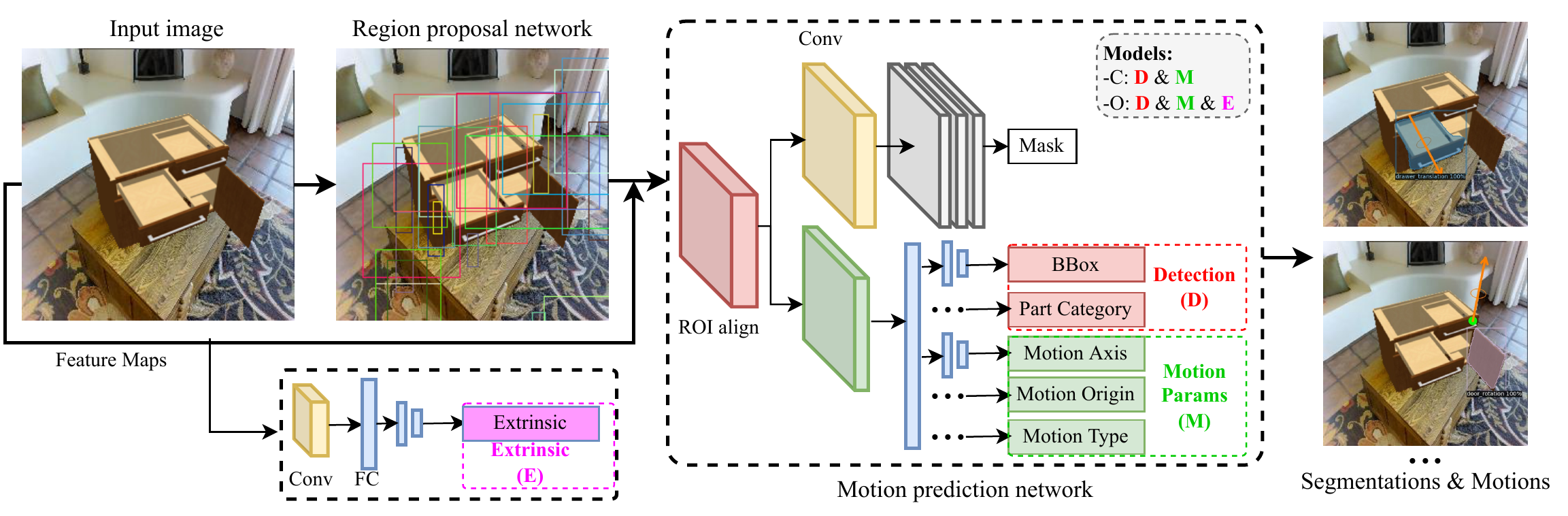}
\caption{
Illustration of the network structure for our \opdnetcc and \opdnetocs architectures.
We leverage a MaskRCNN backbone to detect openable parts. Additional heads are trained to predict motion parameters for each part.
}
\label{fig:network_st}
\end{figure*}

\section{Approach}

To address \ourtasklong, we leverage a instance segmentation network to identify openable parts by treating each part as an `object instance'.
Specifically, we use Mask-RCNN~\cite{he2017mask} and add additional heads for predicting the motion parameters.
Mask-RCNN receives an image and uses a backbone network for feature extraction and a region proposal network (RPN) to propose regions of interest (ROI) which are fed into branches that refine the bounding box and predict the mask and category label.
By training Mask-RCNN on our \ourdataset, we can detect and segment openable parts.
We attach additional branches to the output of the \texttt{RoiAlign} module to predict motion parameters. 
We consider two variants: i) \opdnetcc which directly predicts the motion parameters in the camera coordinate; and ii) \opdnetoc which predicts the extrinsic parameter (object pose in our single-object setting) and the motion parameters in the world coordinate (canonical object coordinate).
\Cref{fig:network_st} shows the overall structure of our approach.
For all models we use a cross-entropy loss for categorical prediction and a smooth L1 loss with $\beta=1$ when regressing real values (see supplement).

\mypara{\opdnetcc.}
For \opdnetcc, we add separate fully-connected layers to the \texttt{RoiAlign} module to directly predict the motion parameters. 
The original MaskRCNN branches predict the part label $l_i$ and part bounding box delta $\delta_i$ for each part $p_i$ in the box head.
The box delta $\delta_i$ is combined with the box proposal from the RPN module to calculate the final output bounding box $b_i$.
We add additional branches to predict the motion parameters $\phi_i = [c_i, a_i, o_i]$ (motion category, motion axis, motion origin) in the camera coordinate frame.
We use the smooth L1 loss for the motion origin and motion axis, and the cross-entropy loss for the part joint type.
Note that we only apply the motion origin loss for revolute joints, since the motion origin is not meaningful for prismatic joints.

\mypara{\opdnetoc.}
For \opdnetoc, we add additional layers to predict the 6 DoF object pose parameters so that we can establish an object coordinate frame within which to predict motion axes and origins for openable parts.
Following prior work~\cite{wang2019normalized}, we define the object coordinate frame to have a consistent up and front orientation.
The motivation is that motion origins and axes in object coordinates are more consistent than in camera coordinates, given that the same articulation can be observed from different camera viewpoints.
We are only dealing with a single object per image and consistent poses are available for each annotated object, so predicting the object pose is equivalent to predicting the extrinsic parameters of the camera pose.

We regress these object pose parameters directly from the image features (see \Cref{fig:network_st}).
We add convolution layers and fully-connected layers to the backbone module and use the image features to regress the extrinsic parameters.  
For training \opdnetoc, we use the same loss function for motion parameters as \opdnetcc, but with predicted and ground-truth motion axes and origins in object coordinates.
For the extrinsic parameters, we treat the matrix as a vector of length 12 (9 for rotation, 3 for translation) and use the smooth L1 loss.

\mypara{Implementation details.}
We use Detectron2~\cite{wu2019detectron2} to implement our architecture.
We initialize the network using weights from a ResNet50 pretrained on ImageNet, and train the network using SGD. 
Unless otherwise specified, we use a learning rate of $0.001$ with linear warmup for $500$ steps and then decaying the learning rate by $0.1$ at $0.6$ and $0.8$ of the total number of steps, following the recommended learning rate schedule for Detectron2.

We first train our model only on the detection and segmentation task with a learning rate of $0.0005$ for $30000$ steps.
Then we pick the best weights for RGB, depth, RGBD independently.
Because our \opdnetcc and \opdnetoc have the same structure for detection and segmentation, we load the weights from the best detection and segmentation models and fully train our network with all losses.
In both \opdnetcc and \opdnetoc we use the ratios $[1, 8, 8]$ for the weights of the individual motion loss components: motion category loss, motion axis loss and motion origin loss respectively.
In \opdnetoc we use $15$ as the weight for the object pose loss.

During training we employ image space data augmentation to avoid overfitting (random flip, random brightness, and random contrast).
During inference, we use a greedy non-maximum suppression (NMS) with IoU threshold $0.5$ and choose the predicted bounding box with highest score.
We use a confidence threshold of $0.05$, and allow for $100$ maximum part detections per image.

\begin{table*}[t]
\caption{Evaluation of openable part detection and part motion parameter estimation on the \ourdatacad test set. The \randmot and \mostfreq baselines use detections and extrinsic parameters from \opdnetocs. Both variants of \method outperform baselines and prior work especially for RGB-only inputs and on metrics accounting for part motion parameter estimation.}
\resizebox{\linewidth}{!}{
\begin{tabular}{@{} ll rrrr rrr @{}}
\toprule
& & \multicolumn{4}{c}{Part-averaged mAP $\% \uparrow$} & \multicolumn{3}{c}{Motion-averaged mAP $\% \uparrow$} \\
\cmidrule(l{0pt}r{2pt}){3-6} \cmidrule(l{2pt}r{0pt}){7-9}
Input & Model & \partdet & +\mtype & +\mtype{}\axis & +\mtype{}\axis{}\orig & \motiondet & +\mtype{}\axis & +\mtype{}\axis{}\orig \\
\midrule
\multirow{5}{*}{RGBD}
& \randmot & 5.0 & 1.3 & 0.2 & 0.1 & 6.2 & 0.7 & 0.3 \\
& \mostfreq & 69.4 & 66.1 & 49.2 & 27.8 & 73.6 & 61.6 & 38.8 \\
\cmidrule(l{0pt}r{0pt}){2-9}
& \opdnetcc & \textbf{69.2}$\pm$0.26 &	\textbf{67.3}$\pm$0.25 &	42.6$\pm$0.25 &	40.9$\pm$0.31 &	\textbf{75.3}$\pm$0.09 &	55.3$\pm$0.23 &	53.9$\pm$0.20  \\
& \opdnetocs & 68.5$\pm$0.31 &	66.6$\pm$0.38 &	\textbf{52.6}$\pm$0.26 &	\textbf{49.0}$\pm$0.21 &	74.9$\pm$0.13 &	\textbf{65.3}$\pm$0.27 &	\textbf{62.9}$\pm$0.24 \\
\midrule
\multirow{2}{*}{D (PC)}
& \ancsh~\cite{li2020category} & 2.7 & 2.7 & 2.3 & 2.1 & 3.9 & 3.1 & 2.8 \\
& \pnetopd & 20.4 & 19.3 & 14.0 & 13.6 & 22.0 & 18.1 & 17.6 \\
\cmidrule(l{0pt}r{0pt}){2-9}
\multirow{3}{*}{D}
& \opdnetcc & \textbf{68.2}$\pm$0.39 &	\textbf{66.5}$\pm$0.42 &	41.0$\pm$0.21 &	39.2$\pm$0.20 &	\textbf{73.0}$\pm$0.09 &	53.0$\pm$0.18 &	51.5$\pm$0.19  \\
& \opdnetocs & 67.3$\pm$0.43 &	65.6$\pm$0.28 &	\textbf{51.2}$\pm$0.44 &	\textbf{47.7}$\pm$0.24 &	72.2$\pm$0.14 &	\textbf{62.3}$\pm$0.25 &	\textbf{60.0}$\pm$0.23\\
\midrule
\multirow{3}{*}{RGB}
& \opdnetcc & \textbf{67.4}$\pm$0.26 &	\textbf{66.2}$\pm$0.18 &	40.9$\pm$0.21 &	38.0$\pm$0.19 &	\textbf{75.0}$\pm$0.14 &	53.4$\pm$0.20 &	51.4$\pm$0.18 \\
& \opdnetocs & 66.6$\pm$0.28 &	65.5$\pm$0.27 &	\textbf{50.7}$\pm$0.23 &	\textbf{46.9}$\pm$0.26 &	74.5$\pm$0.26 &	\textbf{63.8}$\pm$0.32 &	\textbf{61.5}$\pm$0.26\\
\bottomrule
\end{tabular}
}
\label{tab:results-cad-test}
\end{table*}

\section{Experiments}

\subsection{Metrics}

\mypara{Part detection.}
To evaluate part detection and segmentation we use standard object detection and segmentation metrics over the part category, as implemented for MSCOCO~\cite{lin2014microsoft}.
In the main paper, we report mAP@IoU=$0.5$ for the predicted part label and 2D bounding box (\partdet).

\mypara{Part motion.}
To evaluate motion parameter estimation and understand the influence of the motion type we compute detection metrics over the motion type (motion-averaged mAP).
This is in contrast with part-averaged mAP which is over the part category.
A `match' for motion-averaged mAP considers the predicted motion type and the 2D bounding box (\motiondet), and error thresholds for the motion axis and motion origin.
We set the thresholds to $10^\circ$ for axis error and $0.25$ of the object's diagonal length for the origin distance (predicted origin to GT axis line) in all experiments.
Motion parameters are evaluated in the camera coordinate frame.
Motion origins for translation joints are all considered to match.
We report several variants of part-averaged mAP: \partdet, \partdet with motion type matched (+\mtype), \partdet with motion type and motion axis matched (+\mtype{}\axis{}), and \partdet with motion type, motion axis and motion origin matched (+\mtype{}\axis{}\orig).
Motion-averaged mAP has the same variants with \motiondet instead of \partdet.
When evaluating motion parameters, prior work~\cite{wang2019shape2motion,li2020category} has focused on the average error only for correctly detected parts.
In contrast, our metrics incorporate whether the part was successfully detected or not (in addition to error thresholds).

\subsection{Baselines}

As noted by \citet{li2020category}, knowing the canonical object coordinates can assist with predicting the motion axes and origin.
This is because the motion axes are often one of the $\{x,y,z\}$ axes in object coordinates.
Similarly, the motion origin for revolute joints is often at the edge of the object.
So we design baselines that select (randomly or using the most frequent heuristic) from the three axes and the edges and corners of the normalized object bounding box.
These baselines rely on known canonical object coordinates, so we take the part and object pose predictions from \opdnetocs.
For origin prediction we use ground truth object sizes to convert from normalized to canonical coordinates, so these are strong baselines with access to additional information not available to \method.

\mypara{\randmot.}
A lower bound on performance that randomly predicts the part type, motion type, motion axis and motion origin.
The motion origin is picked randomly from 19 points in the corner, edge center, face center and center of the object bounding box.
The motion axis is picked randomly from three axis-aligned directions in the canonical object coordinates.

\mypara{\mostfreq.}
Selects the most frequent motion type, axis, and origin in object coordinates conditioned on the predicted part category (statistics from train set).

\mypara{\ancsh.}
\citet{li2020category}'s ANCSH predicts motion parameters (motion axis, motion origin) and part segmentation for single-view point clouds.
This work assumes a fixed kinematic chain for all objects in a category.
In other words, the number of parts, part labels, and part motion types are given as input.
We re-implement ANCSH in PyTorch matching the results reported by the authors.
Since ANCSH requires a fixed kinematic chain, we choose the most frequent kinematic chain: objects with one rotating door part (243 out of 683 total objects in \ourdatacad).
We train ANCSH on this `one-door dataset'.
The main evaluation is on the complete validation and test set (same as other baselines and approaches).
The supplement provides additional comparisons on only one-door objects.

\mypara{\pnetopd.}
Baseline using a PointNet++ backbone to predict instance segmentation and motion parameters directly from an input point cloud.
This baseline predicts the part category, part instance id, and motion parameters for each point.
This architecture operates on a fixed number of parts so we train on objects with less than 5 parts.
See the supplement for details on the architecture.

\begin{table}[ht]
\caption{
Results on the \ourdatareal test set. Overall, the task is more challenging on real objects.
\opdnetocs has the highest performance across most metrics, and in particular for motion-averaged metrics including motion estimation (+\mtype{}\axis{}\orig).
}
\centering
{\footnotesize
\resizebox{\linewidth}{!}{
\begin{tabular}{@{} ll rrrr rrr @{}}
\toprule
& & \multicolumn{4}{c}{Part-averaged mAP $\% \uparrow$} & \multicolumn{3}{c}{Motion-averaged mAP $\% \uparrow$} \\
\cmidrule(l{0pt}r{2pt}){3-6} \cmidrule(l{2pt}r{0pt}){7-9}
Input & Model & \partdet & +\mtype & +\mtype{}\axis & +\mtype{}\axis{}\orig & \motiondet & +\mtype{}\axis & +\mtype{}\axis{}\orig \\
\midrule
\multirow{4}{*}{RGBD}
& \randmot & 5.3 & 1.4 & 0.1 & 0.1 & 7.1 & 0.5 & 0.3 \\
& \mostfreq & 56.6 & 54.6 & 34.0 & 21.6 & 71.6 & 50.4 & 32.2 \\
\cmidrule(l{0pt}r{0pt}){2-9}
& \opdnetcc & 54.7 & 53.3 & 21.8 & 21.3 & \textbf{73.4} & 32.3 & 31.5 \\
& \opdnetocs & \textbf{56.6} & \textbf{54.3} & \textbf{33.8} & \textbf{32.4} & 73.3 & \textbf{50.0} & \textbf{48.1}  \\
\midrule
\multirow{3}{*}{D}
& \pnetopd & 15.4 & 15.3 & 12.1 & 11.5 & 22.8 & 17.9 & 17.0 \\
& \opdnetcc & 49.4 & 46.6 & 12.2 & 11.6 & \textbf{61.1} & 17.7 & 17.0 \\
& \opdnetocs & \textbf{49.2} & \textbf{47.3} & \textbf{18.1} & \textbf{16.1} & 60.3 & \textbf{26.7} & \textbf{23.9} \\
\midrule
\multirow{3}{*}{RGB}
& \opdnetcc & \textbf{58.0} & \textbf{57.0} & 22.2 & 21.3 & 73.6 & 32.6 & 31.4 \\
& \opdnetocs & 57.8 & 56.4 & \textbf{33.0} & \textbf{30.8} & \textbf{74.0} & \textbf{48.7} & \textbf{45.7} \\
\bottomrule
\end{tabular}
}
}
\label{tab:results-real-test}
\end{table}

\subsection{Results}

\mypara{Model comparisons.} \Cref{tab:results-cad-test} shows the performance of the baselines and the two variants of \method.
We can make a number of observations.
The \randmot baseline performs quite poorly for both part detection and part motion estimation metrics on all input scenarios, indicating the challenge of detecting openable parts and estimating their motion parameters.
The \mostfreq baseline is quite competitive if we only look at part detection and motion type detection (\partdet and \motiondet metrics).
This is not surprising as the \mostfreq baseline leverages detections from \opdnetocs and the most frequent motion heuristic is relatively strong.
However, when we look at precision of motion axis and motion origin predictions (+\mtype{}\axis{}\orig and related metrics) we see that \opdnetocs significantly outperforms simpler baselines, especially for the motion-averaged metrics (higher than 60\% mAP).
It also outperforms the camera-centric \opdnetcc, showing the benefit of object-centric representation relative to camera-centric representation when accurate estimation of motion axes and origins is important.

\mypara{Effect of input modalities.} Comparing different input modalities in \Cref{tab:results-cad-test} we see that overall, methods do best with RGBD, while D (depth, or point cloud) input, and RGB-only input are more challenging.
For the depth input modality we compare against \ancsh~\cite{li2020category} and the \pnetopd baseline.
Note that \ancsh requires separate models for different kinematic chains so it is severely disadvantaged when evaluating on the \ourdatacad dataset that includes objects with varying number of parts and differing motion types.
We observe that \ancsh is outperformed by \pnetopd and both variations of \method.
One of the limitations of \ancsh is that it requires a prespecified kinematic chain (with fixed number of parts, part categories and motion types).
Therefore, we also evaluated \ancsh in a setting constrained to `single door' objects, where it performs more competitively (see supplement).
These observations demonstrate the increased generality of our approach over baselines in terms of handling arbitrary object categories with changing numbers of moving parts and motion types.
Moreover, both \method variants perform well in the RGB-only setting.

\mypara{Error metrics.}
We also compute error metrics as in \citet{li2020category}.
Prior work computes these metrics only for detected parts that match a ground truth part, thus including different number of instances for different methods.
\ancsh and \pnetopd have average motion axis error of $10.36^\circ$ (for $6975$ instances) and $6.25^\circ$ (for $9862$ instances) respectively.
Both have average motion origin error of $0.09$.
Our proposed methods have comparable errors but over more detected parts.
\opdnetcc and \opdnetoc have axis error of $9.06^\circ$ and $6.67^\circ$ for $\sim22000$ instances, and origin error of $0.11$.
These error metrics are restricted only to matched predictions and fail to capture detection performance differences.
The supplement provides more analysis using error metrics.

\mypara{Real-world images.}
We demonstrate that we can apply our method on real-world images to detect openable parts and their motion parameters.
We finetune the RGB and RGBD models trained on \ourdatacad on the \ourdatareal training set.
All hyperparameters are the same except we set the learning rates to 0.001 for RGB and RGBD, and 0.0001 for depth-only (D) models.
\Cref{tab:results-real-test} summarizes performance on our \ourdatareal dataset.
Overall, the task is much more challenging with real data as seen by lower performance across all metrics for all methods.
We see that \opdnetoc has the best performance overall across most metrics, and in particular for metrics that measure motion parameter estimation.

\begin{figure*}
\centering
\setkeys{Gin}{width=\linewidth}
\begin{tabularx}{\textwidth}{Y Y Y Y Y Y Y Y Y}
\toprule

\multicolumn{9}{c}{\ourdatacad}\\
\midrule

\tiny{GT} &
\imgclip{0}{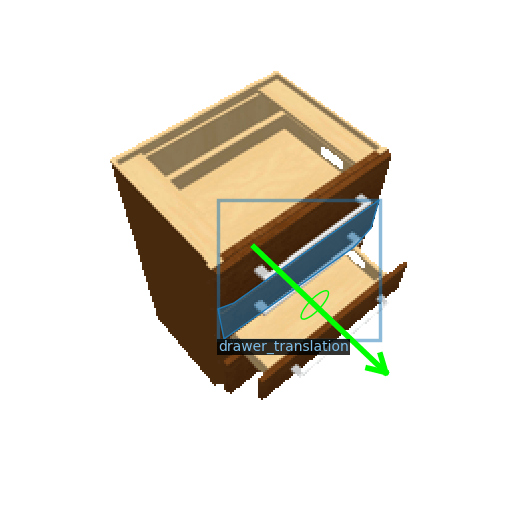} & 
\imgclip{0}{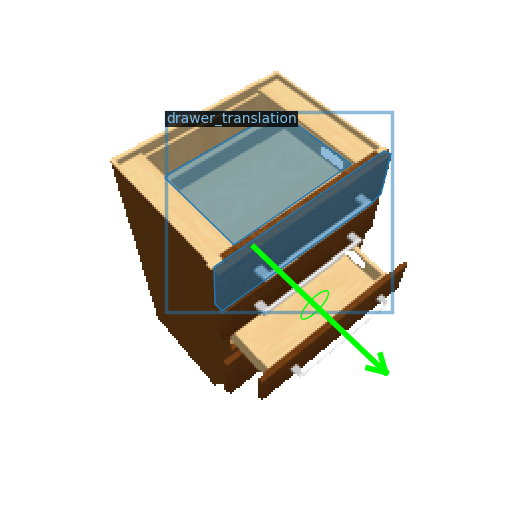} &
\imgclip{0}{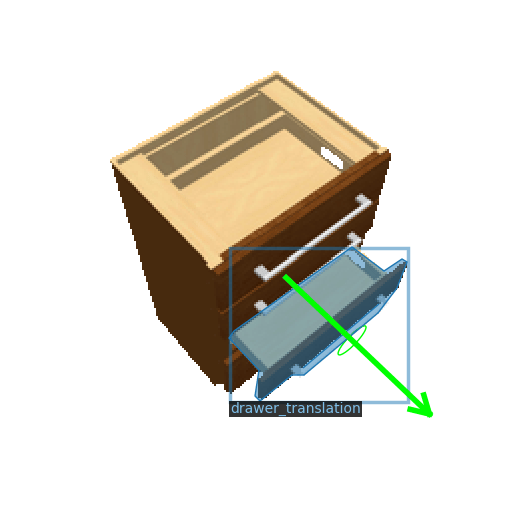} &
\imgclip{0}{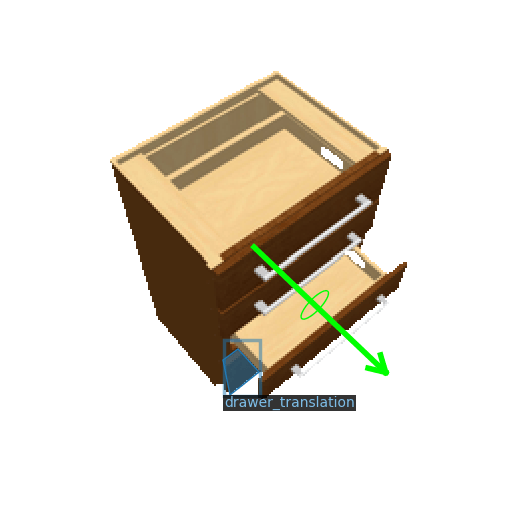} &
\imgclip{0}{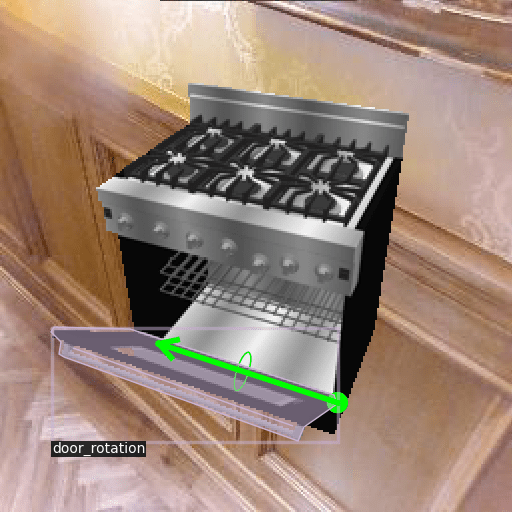} &
\imgclip{0}{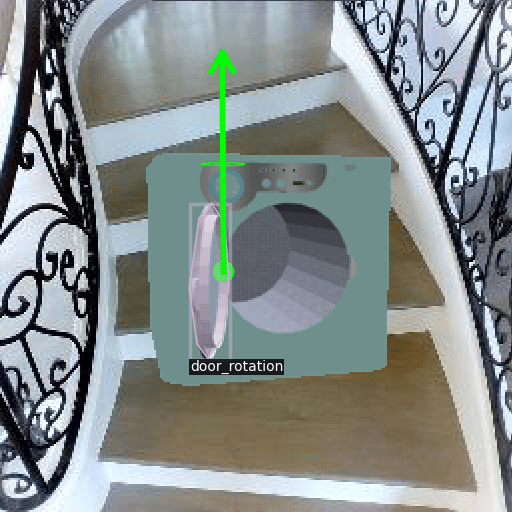} &
\imgclip{0}{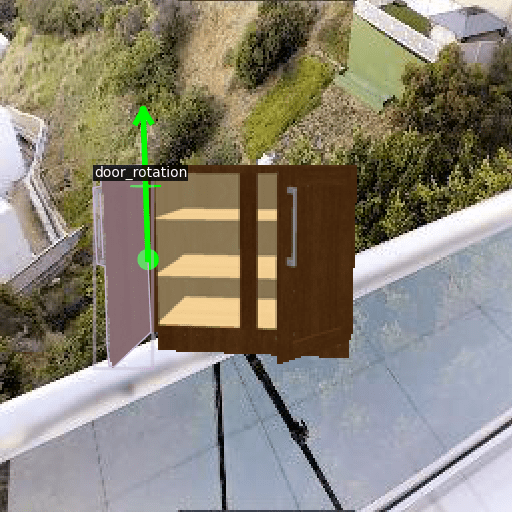} &
\imgclip{0}{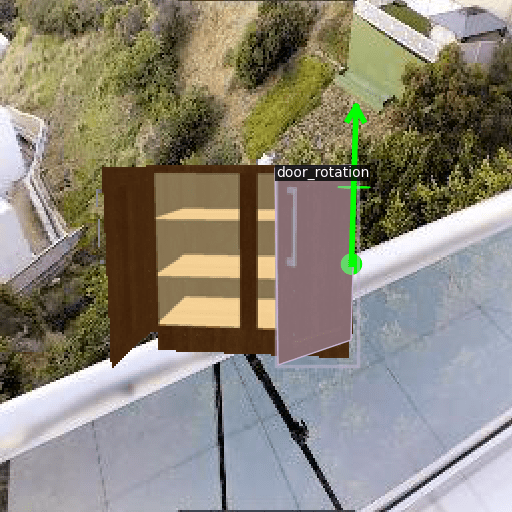}\\

\tiny{\opdnetcc} &
\imgclip{0}{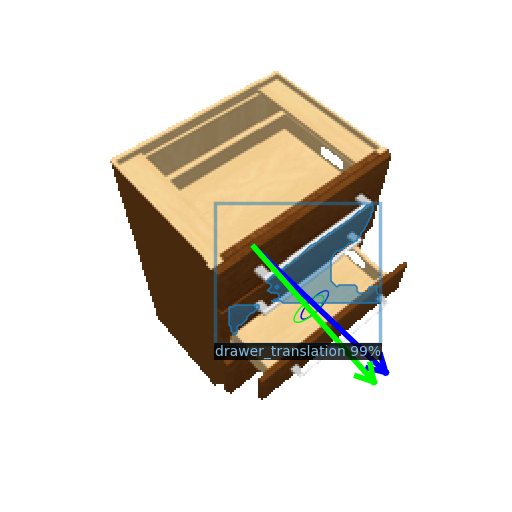} & 
\imgclip{0}{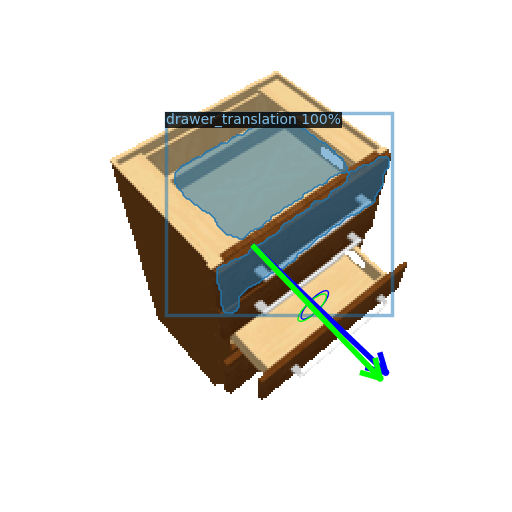} &
\imgclip{0}{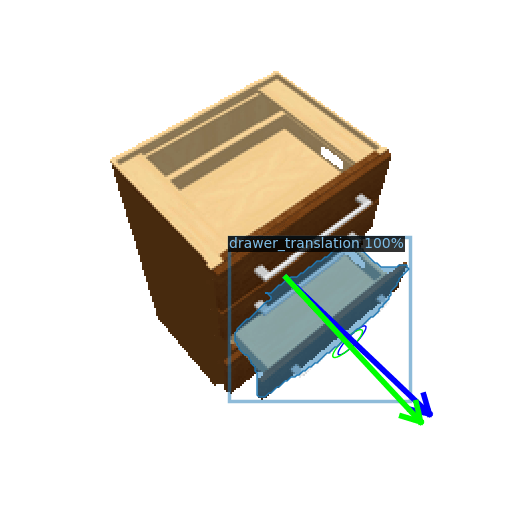} &
\imgclip{0}{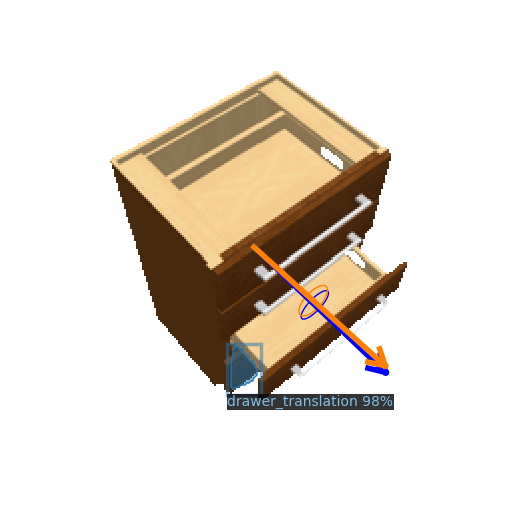} &
\imgclip{0}{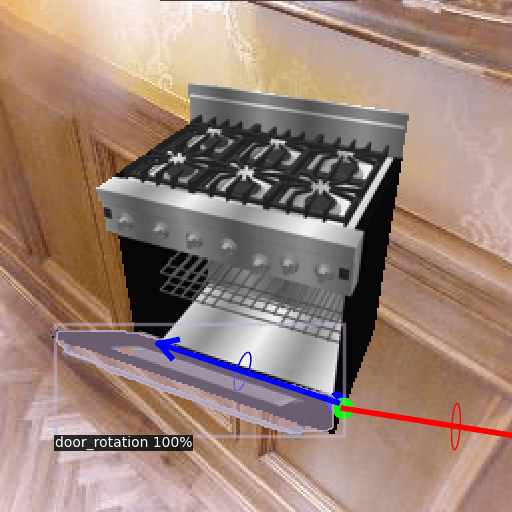} &
\imgclip{0}{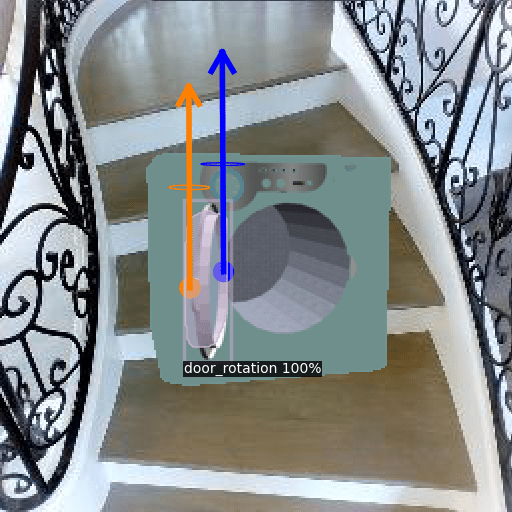} &
\imgclip{0}{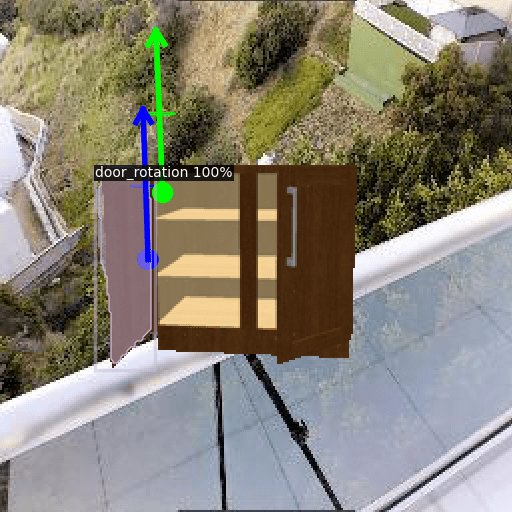} &
\imgclip{0}{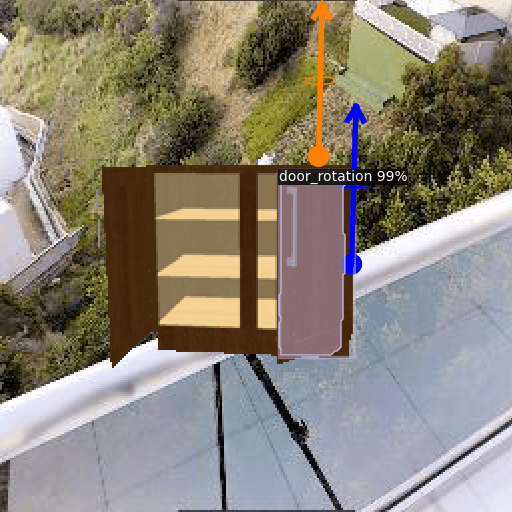} \\

\tiny{\opdnetocs} &
\imgclip{0}{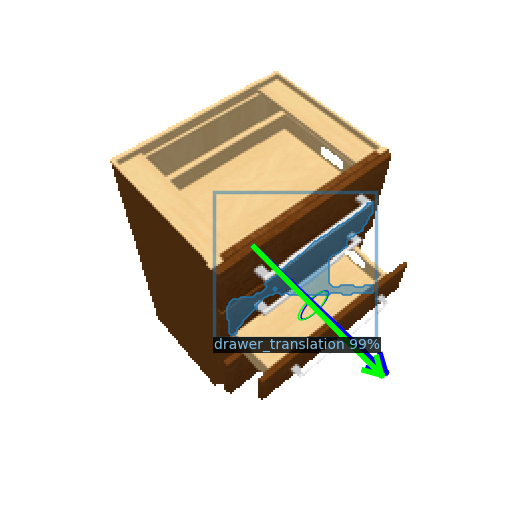} & 
\imgclip{0}{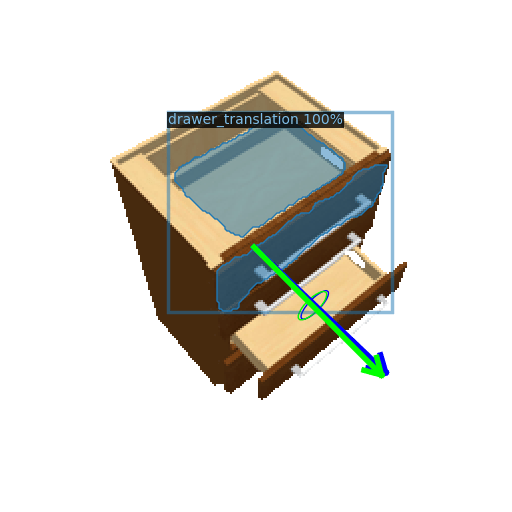} &
\imgclip{0}{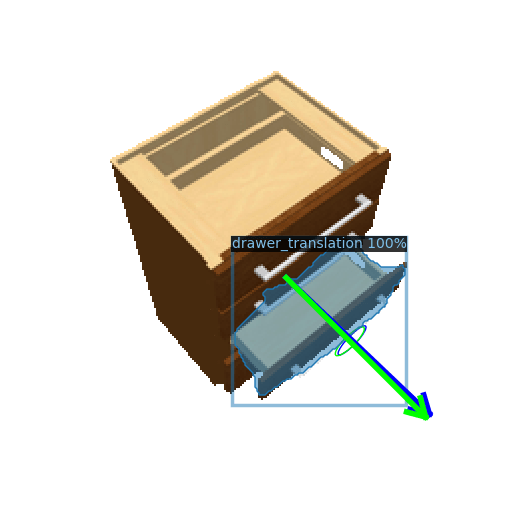} &
\imgclip{0}{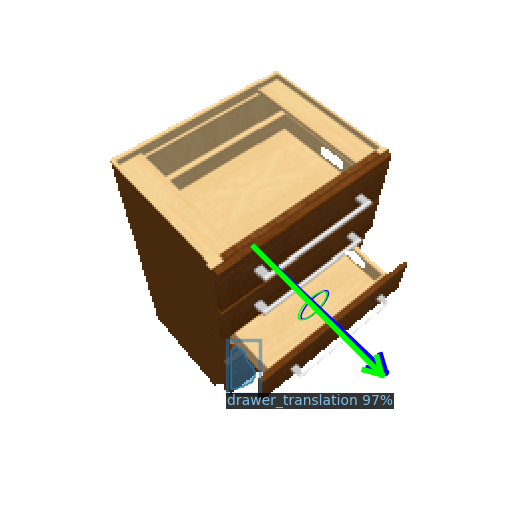} &
\imgclip{0}{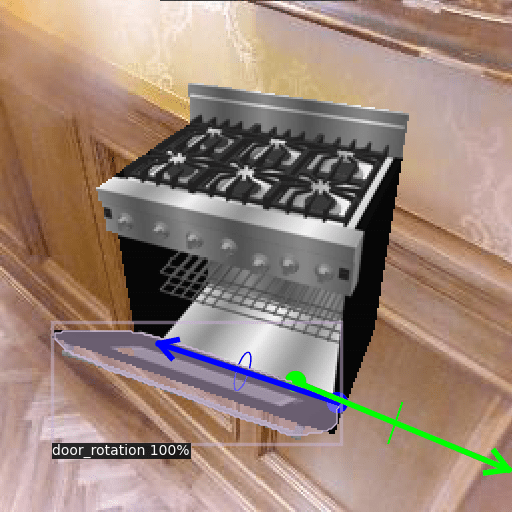} &
\imgclip{0}{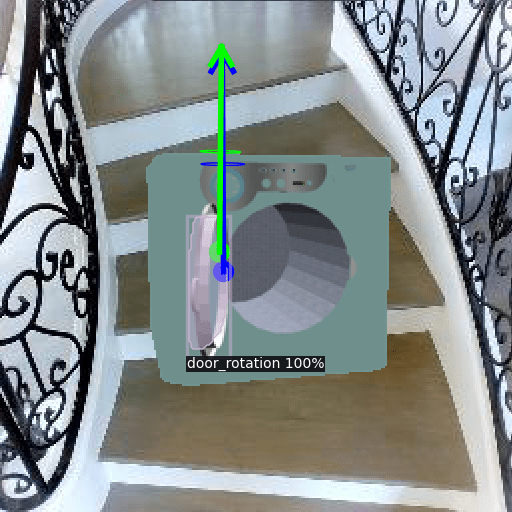} &
\imgclip{0}{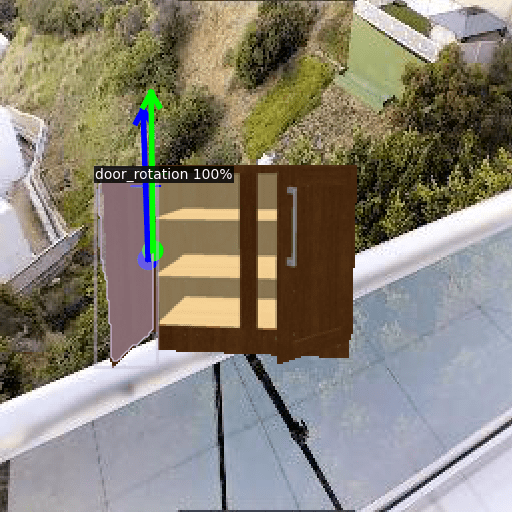} &
\imgclip{0}{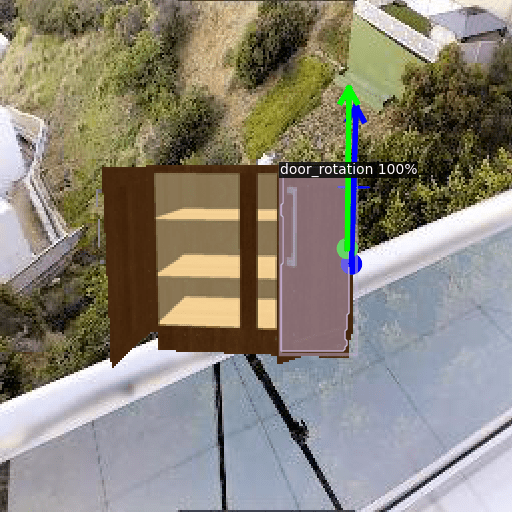} \\

\toprule
\multicolumn{9}{c}{\ourdatareal}\\
\midrule

\tiny{GT} &
\imgclip{0}{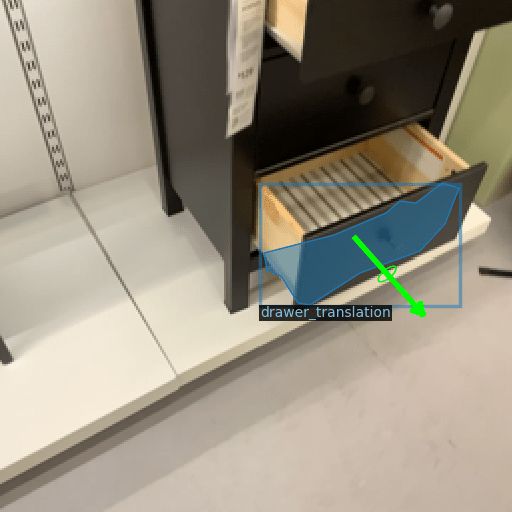} &
\imgclip{0}{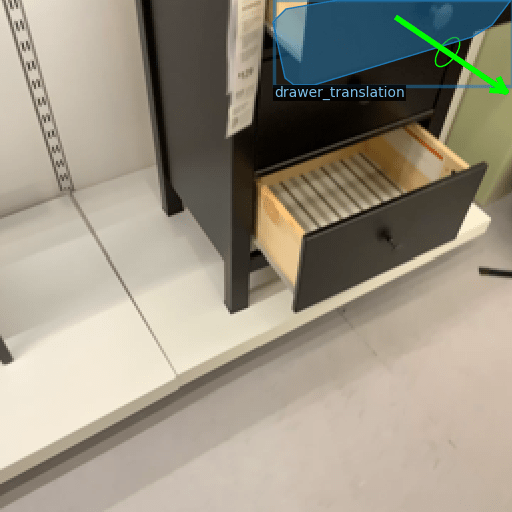} &
\imgclip{0}{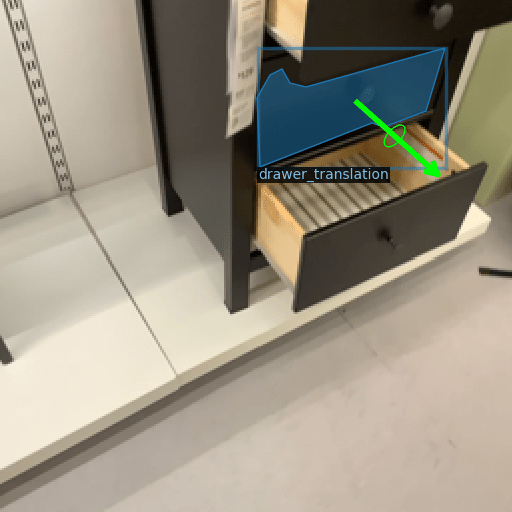} &
\imgclip{0}{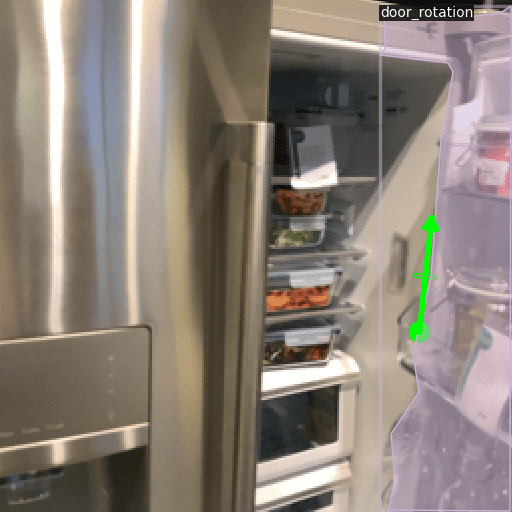} &
\imgclip{0}{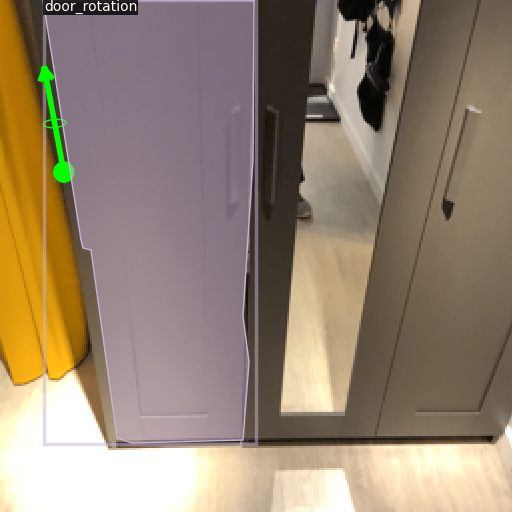} &
\imgclip{0}{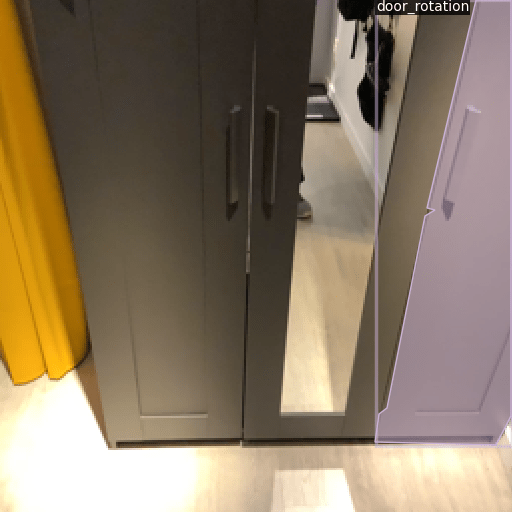} &
\imgclip{0}{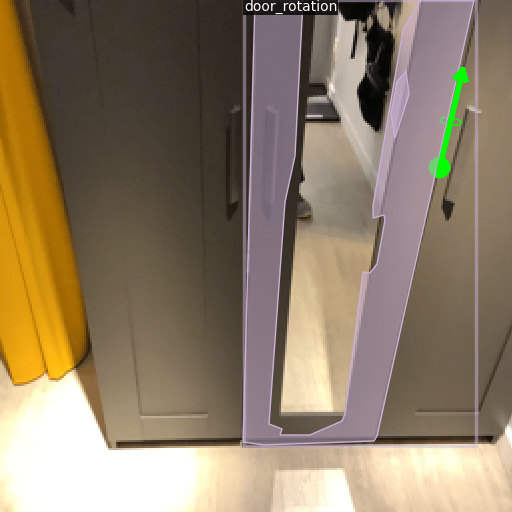} &
\imgclip{0}{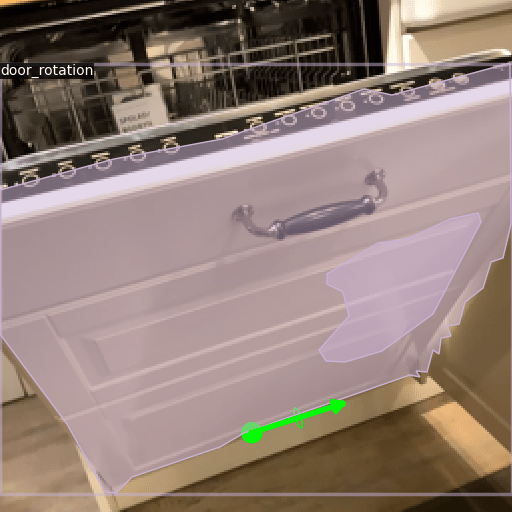} \\

\tiny{\opdnetcc} &
\imgclip{0}{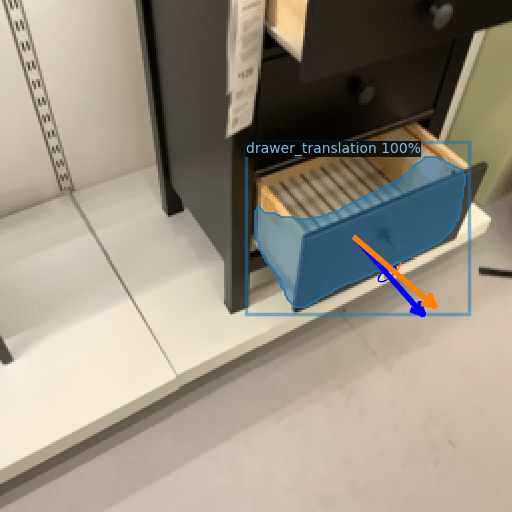} &
\imgclip{0}{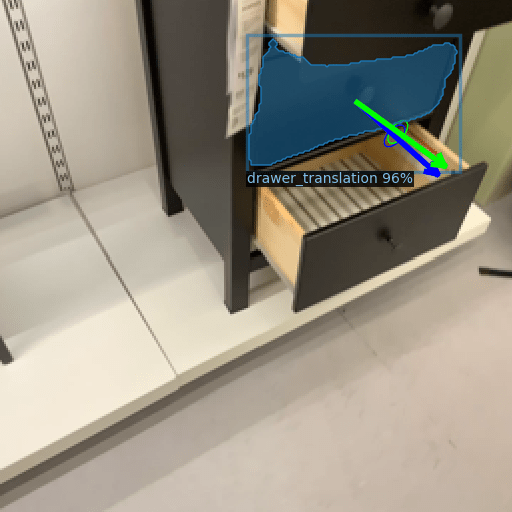} &
\imgclip{0}{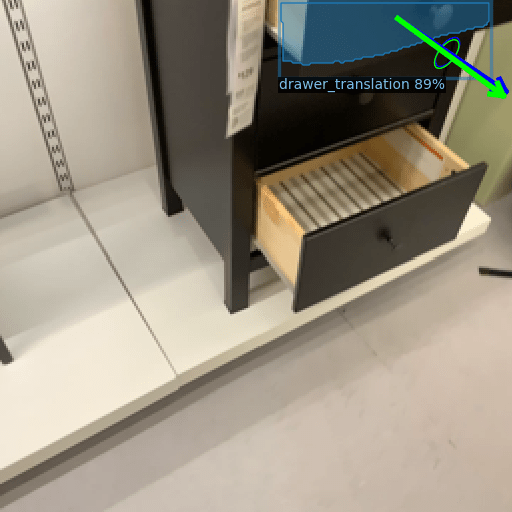} &
\imgclip{0}{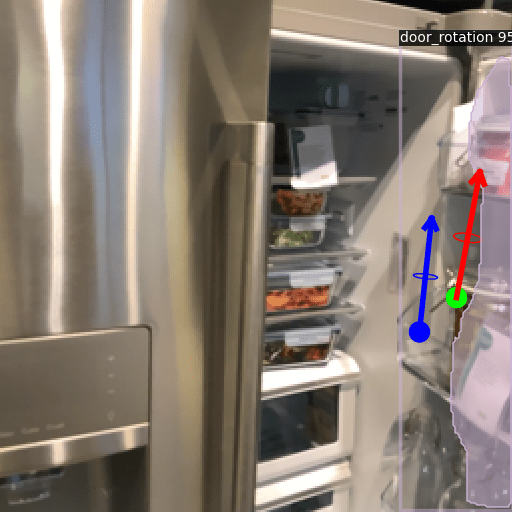} &
\imgclip{0}{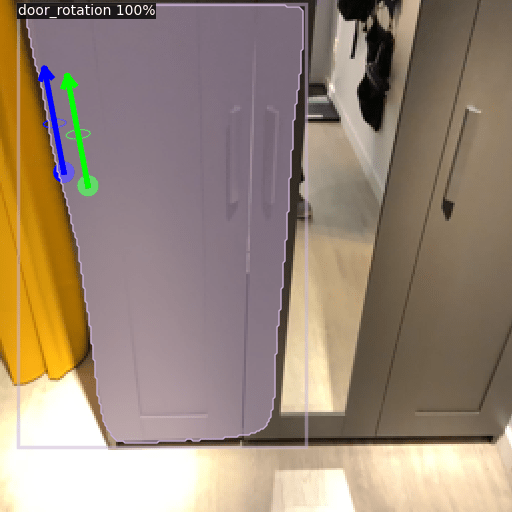} &
Miss &
Miss &
\imgclip{0}{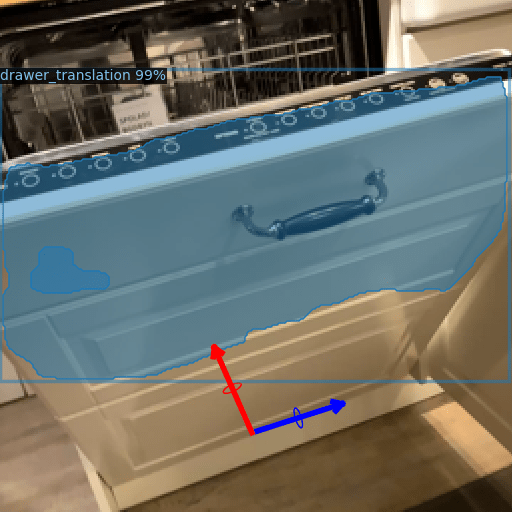} \\

\tiny{\opdnetocs} &
\imgclip{0}{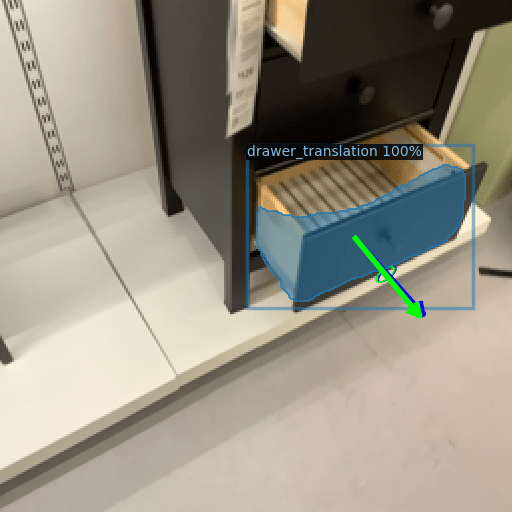} &
\imgclip{0}{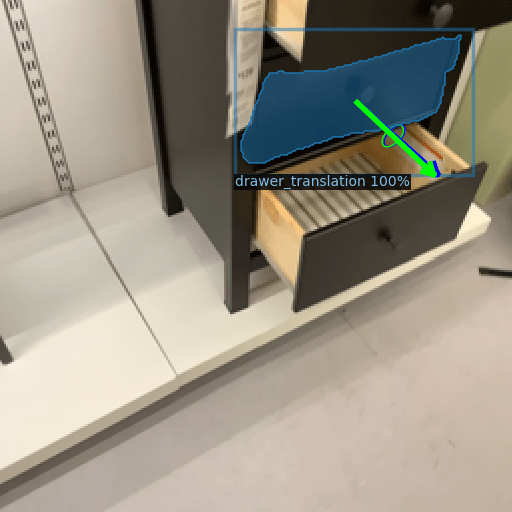} &
\imgclip{0}{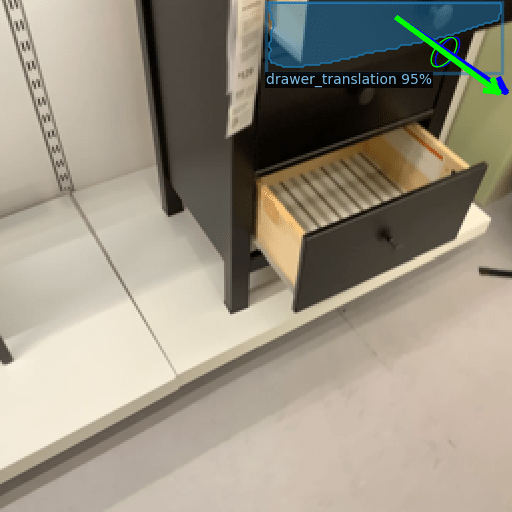} &
\imgclip{0}{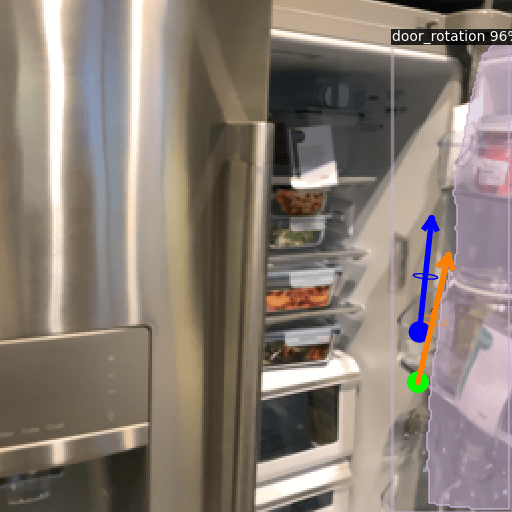} &
\imgclip{0}{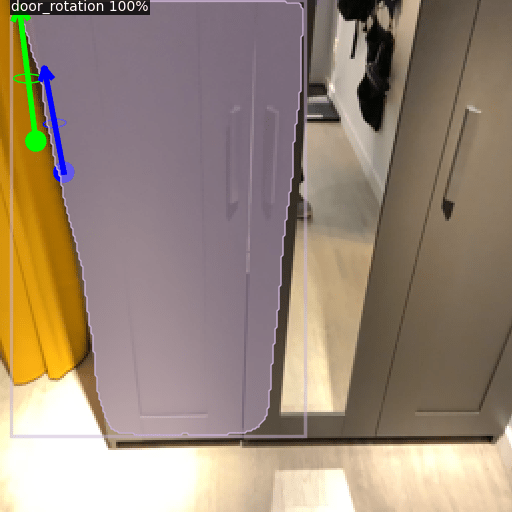} &
\imgclip{0}{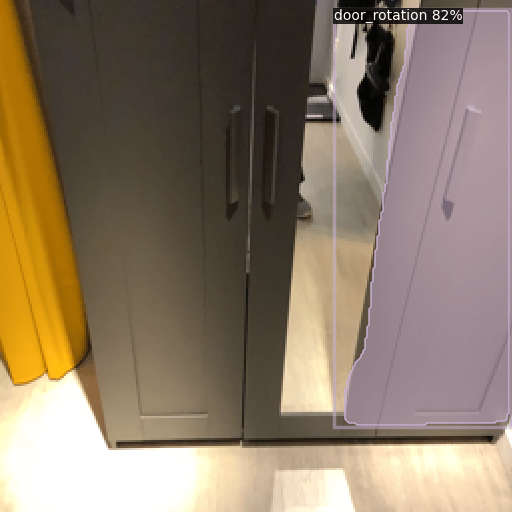} &
\imgclip{0}{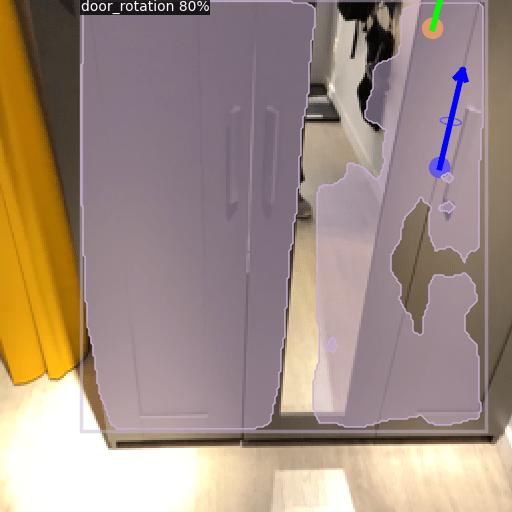} &
\imgclip{0}{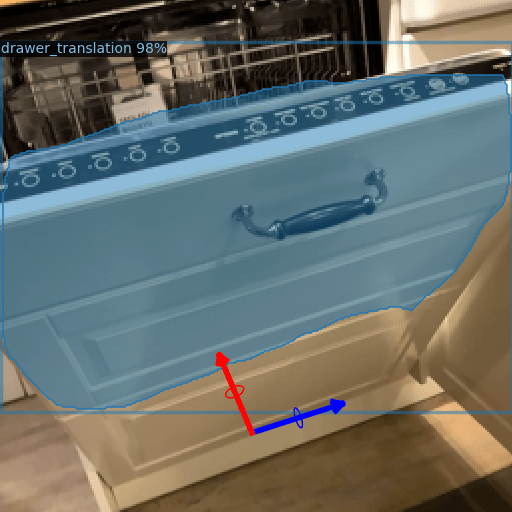} \\

\tiny{GT} &
\imgclip{0}{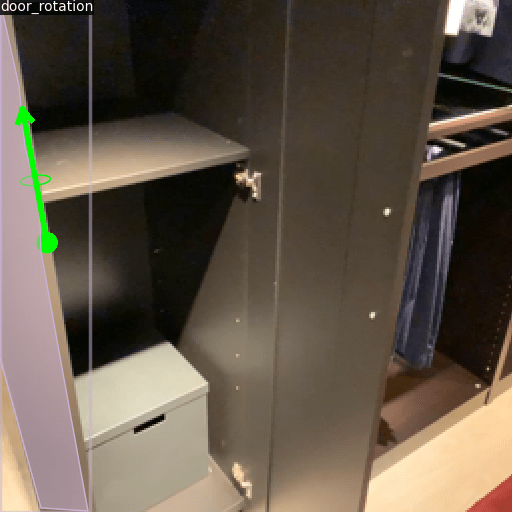} &
\imgclip{0}{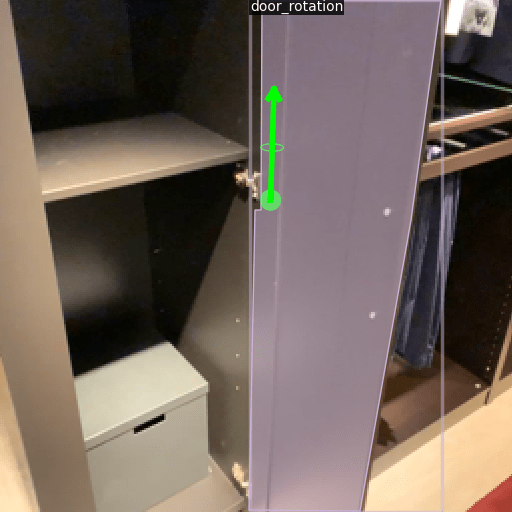} &
\imgclip{0}{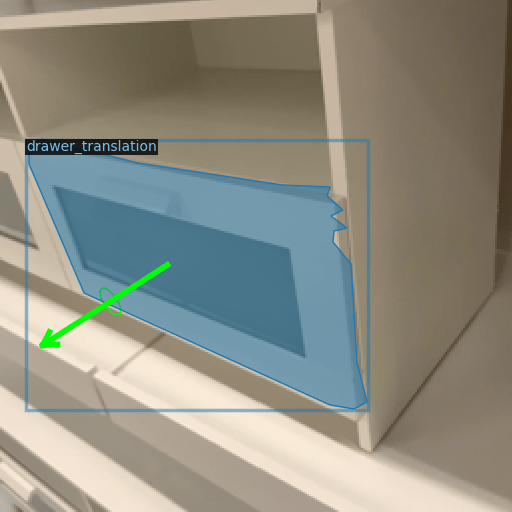} &
\imgclip{0}{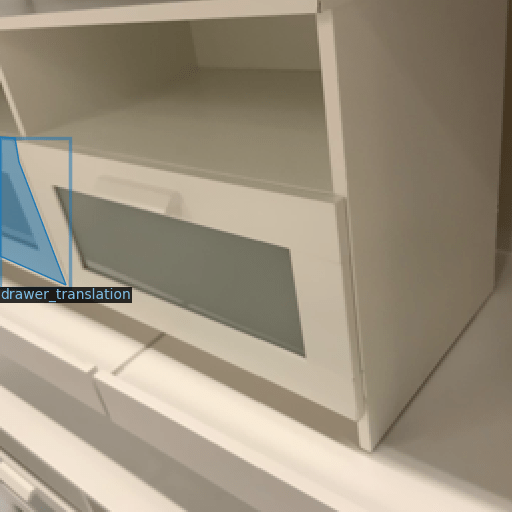} &
\imgclip{0}{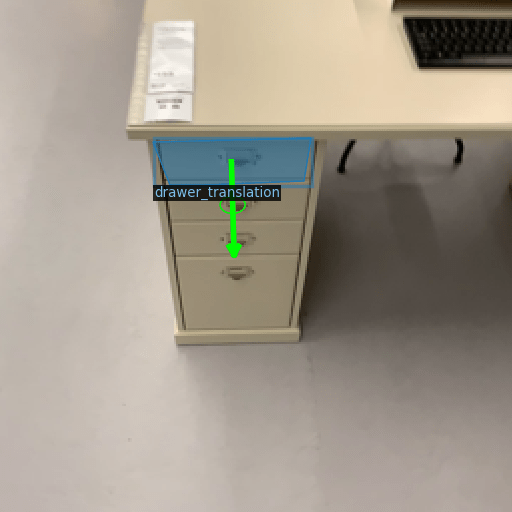} &
\imgclip{0}{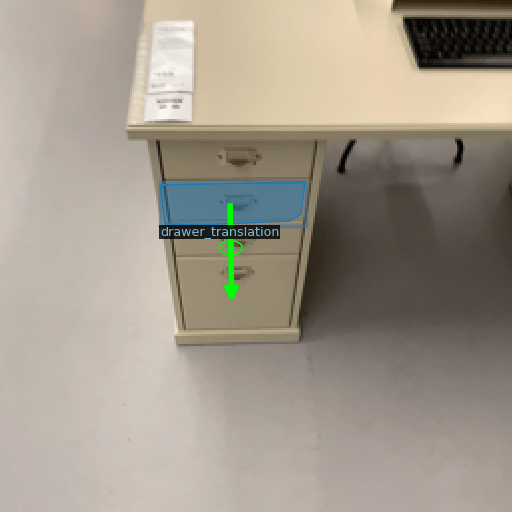} &
\imgclip{0}{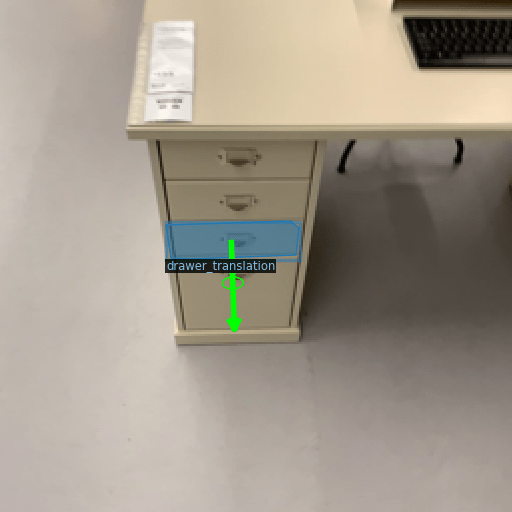} &
\imgclip{0}{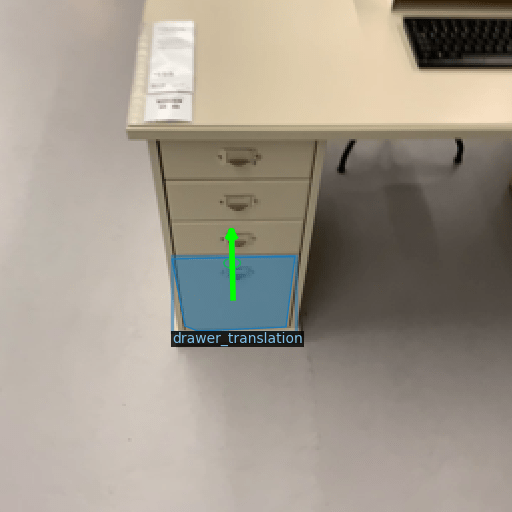} \\

\tiny{\opdnetcc} &
\imgclip{0}{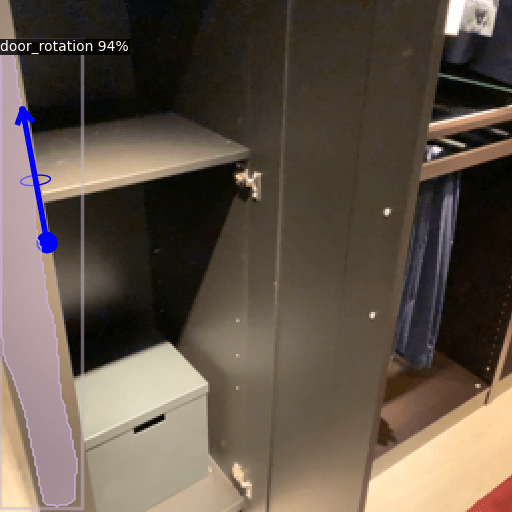} &
\imgclip{0}{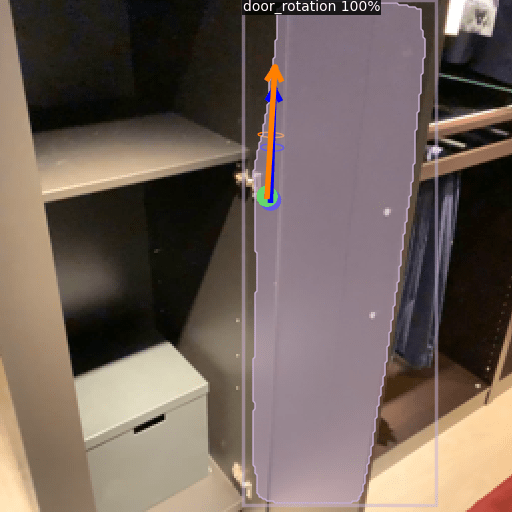} &
\imgclip{0}{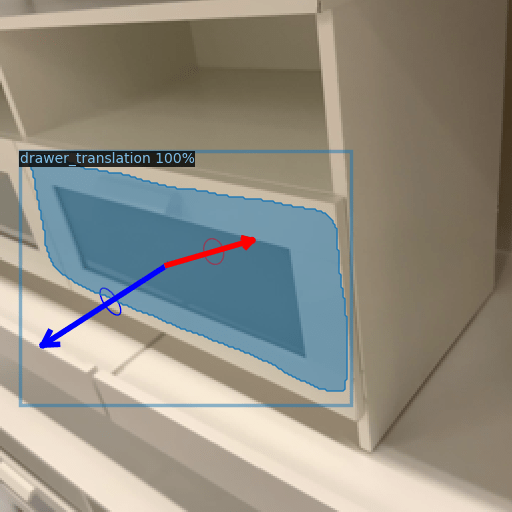} &
Miss &
\imgclip{0}{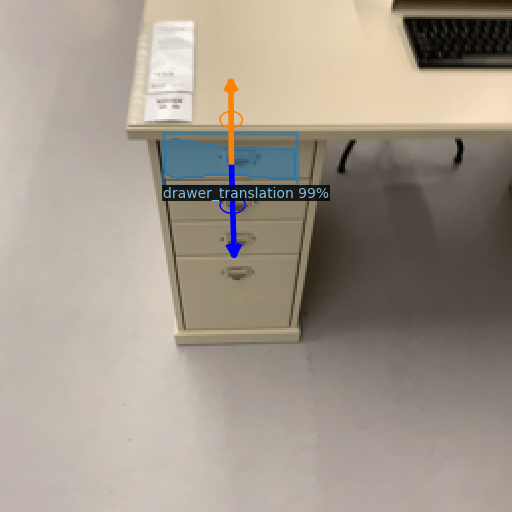} &
\imgclip{0}{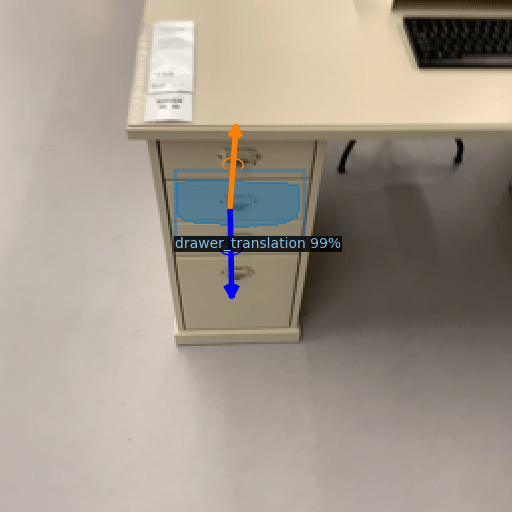} &
\imgclip{0}{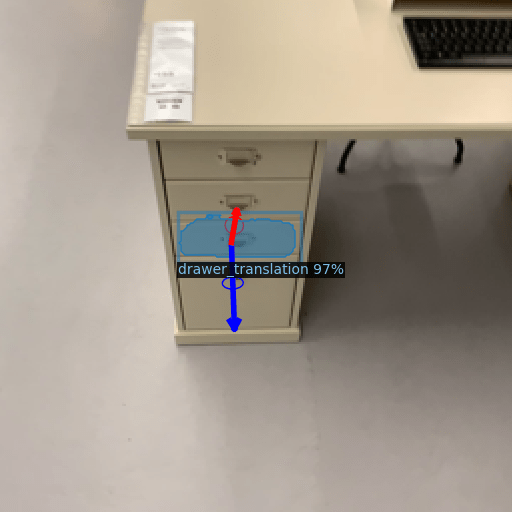} &
\imgclip{0}{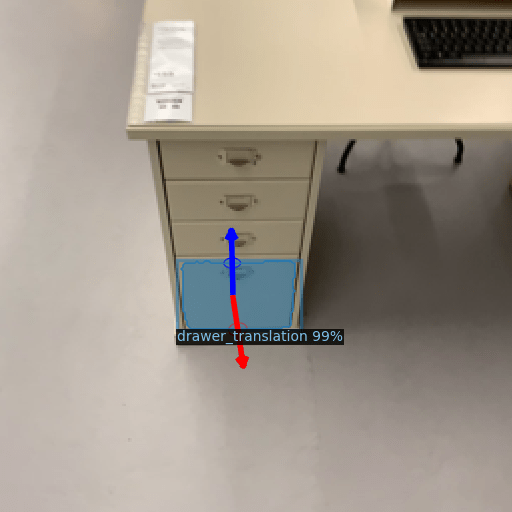} \\

\tiny{\opdnetocs} &
\imgclip{0}{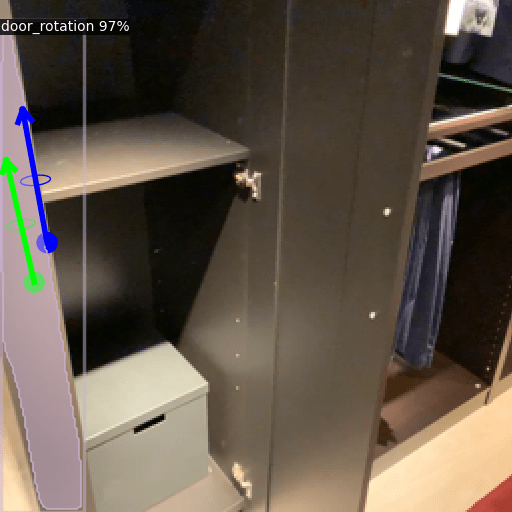} &
\imgclip{0}{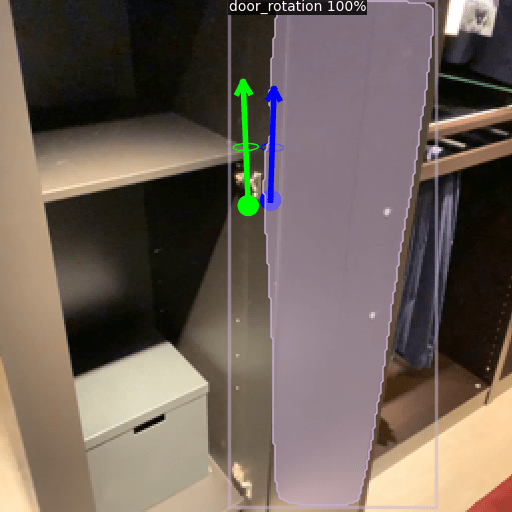} &
\imgclip{0}{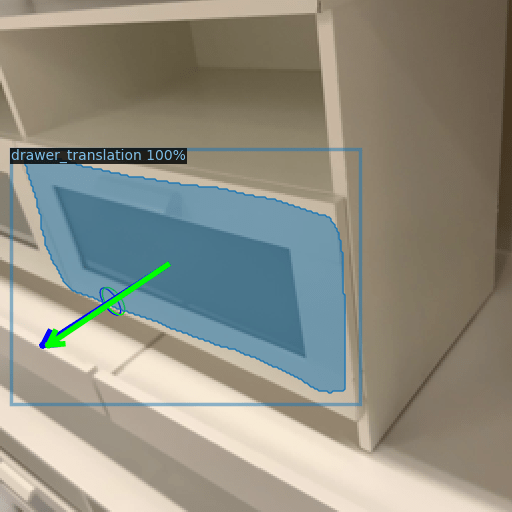} &
\imgclip{0}{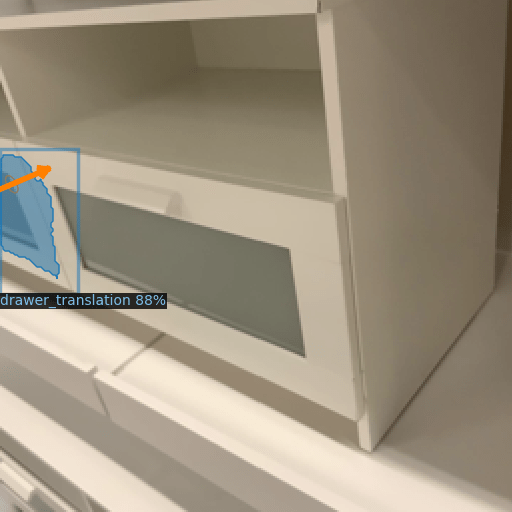} &
\imgclip{0}{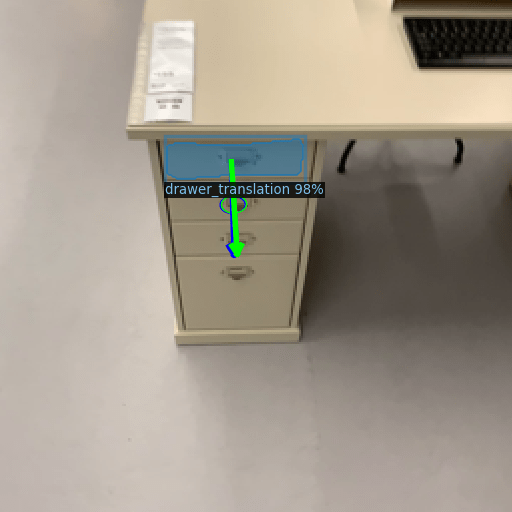} &
\imgclip{0}{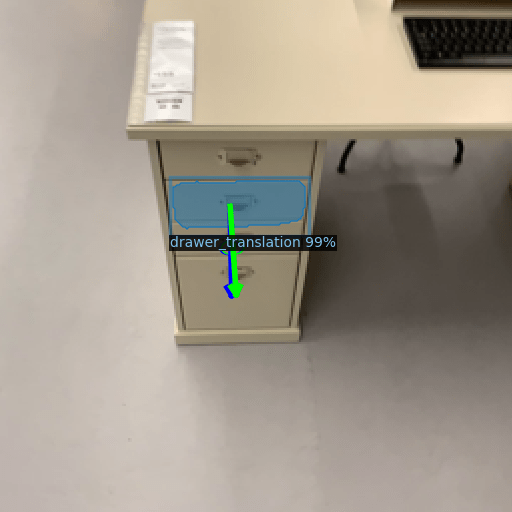} &
\imgclip{0}{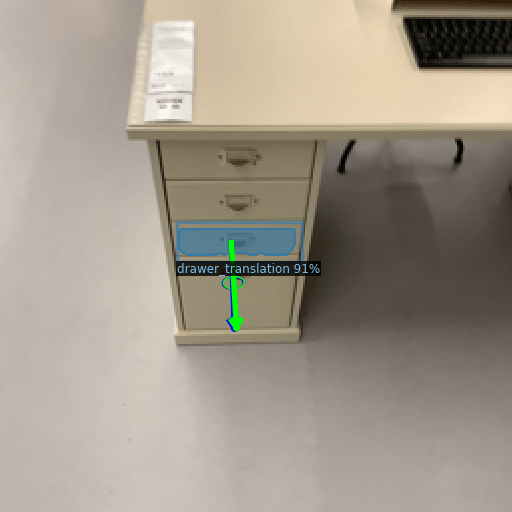} &
\imgclip{0}{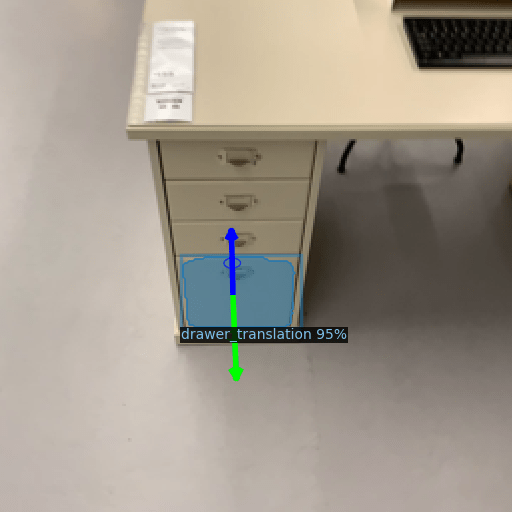} \\

\bottomrule
\end{tabularx}
\caption{Qualitative results from the \ourdatacad and \ourdatareal val sets. The first row in each triplet shows the ground truth (GT) with each openable part mask, and the translational or rotational axis indicated in green. Other rows are predictions using \opdnetcc and \opdnetocs on RGBD inputs. The predicted axis is green if within $5^\circ$ of the ground truth (also shown, in blue), in orange if within $10^\circ$, and in red if more than $10^\circ$. The first few \ourdatacad examples show that translational \drawer openable parts are relatively easy. Rotating \door parts are more challenging (see high error predictions by \opdnetcc in the second row from the top). The \ourdatareal data is more challenging with many high axis error cases and entirely undetected parts (Miss). Particularly hard examples include unusual rotating doors that look like translating drawers (last column in top set), and openable parts only partly visible in the image frame (fourth column).
}
\label{fig:qual-results}
\end{figure*}

\mypara{Qualitative visualizations.} \Cref{fig:qual-results} shows qualitative results from both the \ourdatacad and \ourdatareal datasets.
We see the overall trend that \opdnetocs outperforms \opdnetcc in terms of motion parameter estimation.
We also see that both datasets have hard cases, with \ourdatareal being particularly challenging due to real-world appearance variability and the limited field of view of single-view images resulting in many partially observed parts.

\subsection{Analysis}

\mypara{Are some parts more challenging than others?}
\Cref{tab:results-parts} shows the performance of \opdnetocs broken down by openable part category.
The \drawer parts exhibit translational motion, \lidd parts have rotational motions, and \door parts exhibit both.
We see that \lidd is more challenging than \drawer and \door, with considerably lower part detection mAP and significantly lower part motion estimation performance (+\mtype{}\axis{} and +\mtype{}\axis{}\orig metrics).
This may be caused by fewer \lidd in the training data (89 {\lidd} vs 508 {\door} and 363 {\drawer} parts).

\begin{table*}[t]
\caption{Analysis of performance given ground truth part category, 2D bounding box, and object pose. Results for \opdnetoc are on the \ourdatacad val set.}
\resizebox{\linewidth}{!}{
\begin{tabular}{@{} l rrrr rrr @{}}
\toprule
& \multicolumn{4}{c}{Part-averaged mAP $\% \uparrow$} & \multicolumn{3}{c}{Motion-averaged mAP $\% \uparrow$} \\
\cmidrule(l{0pt}r{2pt}){2-5} \cmidrule(l{2pt}r{0pt}){6-8}
 & \partdet & +\mtype & +\mtype{}\axis & +\mtype{}\axis{}\orig & \motiondet & +\mtype{}\axis & +\mtype{}\axis{}\orig \\
\midrule
RGBD \opdnetoc & 72.5$\pm$0.34 & 70.6$\pm$0.29 & 51.7$\pm$0.62 & 47.1$\pm$0.59 & 75.4$\pm$0.07 & 61.6$\pm$0.32 & 59.0$\pm$0.32 \\
\gtboxpart & \textbf{99.0}$\pm$0.00 &	\textbf{90.9}$\pm$0.16 & 50.6$\pm$0.36 &	45.4$\pm$0.27 &	\textbf{89.7}$\pm$0.15 &	58.1$\pm$0.32 &	54.7$\pm$0.28 \\
\gtpose & 73.1$\pm$0.10 &	71.0$\pm$0.05 &	60.5$\pm$0.06 &	59.4$\pm$0.05 &	75.2$\pm$0.08 &	67.0$\pm$0.14 &	66.2$\pm$0.09 \\
\gtboxpartpose & \textbf{99.0}$\pm$0.00 &	90.6$\pm$0.37 &	\textbf{65.5}$\pm$0.24 &	\textbf{63.8}$\pm$0.17 &	89.5$\pm$0.19 &	\textbf{73.3}$\pm$0.26 &	\textbf{72.0}$\pm$0.30 \\
\bottomrule
\end{tabular}
}
\label{tab:results-gt}
\end{table*}
\begin{table*}[t]
\caption{
Per-category evaluation of\opdnetocs model on the \ourdatacad val set.
All metrics use part-averaged mAP.
The \drawer openable parts are easiest overall and do not benefit much from depth information.
In contrast \door and in particular \lidd parts are more challenging and do benefit from depth in the input.
}
\resizebox{\linewidth}{!}{\begin{tabular}{@{} l rrrr rrrr rrrr @{}}
\toprule
& \multicolumn{4}{c}{\drawer} & \multicolumn{4}{c}{\door} & \multicolumn{4}{c}{\lidd} \\
\cmidrule(l{0pt}r{2pt}){2-5} \cmidrule(l{2pt}r{2pt}){6-9} \cmidrule(l{2pt}r{0pt}){10-13}
Input & \partdet & +\mtype & +\mtype{}\axis & +\mtype{}\axis{}\orig & \partdet & +\mtype & +\mtype{}\axis & +\mtype{}\axis{}\orig & \partdet & +\mtype & +\mtype{}\axis & +\mtype{}\axis{}\orig \\
\midrule
RGB & \textbf{81.4} & \textbf{80.9} & \textbf{71.5} & \textbf{71.5} & \textbf{86.0} & \textbf{81.1} & 61.6 & 57.0 & 58.7 & 58.4 & 27.7 & 17.8 \\
D & 70.3 & 70.1 & 63.4 & 63.4 & 83.3 & 78.6 & 61.7 & 57.4 & 57.8 & 57.5 & 30.0 & 18.3 \\
RGBD & 71.9 & 71.4 & 65.9 & 65.9 & 85.9 & 79.9 & \textbf{63.7} & \textbf{59.8} & \textbf{62.2} & \textbf{61.8} & \textbf{31.4} & \textbf{19.7} \\
\bottomrule
\end{tabular}
}
\label{tab:results-parts}
\end{table*}

\mypara{How does ground truth part and pose affect motion prediction?}
To understand how part detection influences motion prediction, we use ground truth (GT) for the part label, part 2D bounding box, and object pose with \opdnetocs on the \ourdatacad validation set (see \Cref{tab:results-gt}.  
As expected, when the GT part label and box are provided \partdet is close to 100\%.
Surprisingly, having just the GT part label and box does not improve motion prediction.
The ground truth object pose is more important for predicting the motion axis and origin correctly.
Even with GT pose and 2D part information, the motion prediction is still imperfect.
See the supplement for additional analysis. 

\mypara{Does depth information help?}
\Cref{tab:results-cad-test} shows that depth only (D) models are outperformed by RGB and RGBD models.
Depth information is helpful in conjuction with RGB information as seen by the small performance boost between RGB and RGBD across all metrics.
Depth-only models perform worse than RGB-only models across all motion-averaged metrics.
We suspect that this is because most openable parts have minimal difference in depth values along their edges, and thus color is more helpful than depth for predicting openable part segmentation masks.
\Cref{tab:results-parts} provides some insight.
We see that for the \drawer and \lidd category where detection is overall more challenging, depth information does not help and the motion prediction results are also bad.
For the \door category where detection results are higher depth offers additional information that improves motion parameter estimation (higher +\mtype{}\axis{} and +\mtype{}\axis{}\orig metrics for depth-only (D) and RGBD models).
See the supplement for an additional analysis based on breaking down performance across detection mAP ranges.

\mypara{Can we infer part motion states?}
A discrete notion of `motion state' is often useful (e.g., ``the fridge door is open'' vs ``the fridge door is closed'').
We manually annotate a binary open vs closed motion state and continuous distance or angle offset from the closed state for all objects. 
We then add an MLP to the box head to predict binary motion state under a smooth L1 loss.
We finetune the \opdnetocs model to predict the binary motion state along with part detection and motion parameter estimation.
Finetuning is done for 5000 mini-batches.
To evaluate motion state prediction we compute mAP values for `match' or `no match' of the binary motion state.
Here, we define the mAP over motion type to evaluate the motion state prediction.
Part-averaged mAP of the \opdnetoc model on the \ourdatacad validation set is 62.8\% for RGBD, 62.1\% for RGB and 60.7\% for D inputs.
Motion-averaged mAP values are 62.0\%, 64.2\% and 59.9\% respectively.
These values should be contrasted with \partdet and \motiondet in \Cref{tab:results-cad-test}.
We see that overall we can predict binary part motion states fairly well, though this too is a non-trivial task.
We hypothesize that learning a `threshold' for a binary notion of open vs closed is challenging (e.g., ``fridge door cracked open'').

\subsection{Limitations}

We focused on single objects with openable parts.
The objects are fairly simple household objects without complex kinematic chains or complex motions (e.g., we cannot handle bifold doors).
We also did not consider image inputs with multiple objects, each potentially possessing openable parts.
A simple strategy to address this limitation would be to first detect distinct objects and then apply our approach on each object.
Lastly, we focused on estimating the translation or rotation axis and rotation origin parameters but we do not estimate the range of motion for each part.
This would be required to estimate a full part pose and to track the motion of an openable part from RGB video data.
\section{Conclusion}
We proposed the task of openable part detection and motion parameter estimation for single-view RGB images.
We created a dataset of images from synthetic 3D articulated objects (\ourdatacad) and of real objects reconstructed using RGBD sensors (\ourdatareal).
We used these datasets to systematically study the performance of approaches for the openable part detection and motion estimation task, and investigate what aspects of the task are challenging.
We found that the openable part detection task from RGB images is challenging especially when generalization across object and part categories is important.
Our work is a first step, and there is much potential for future work in better understanding of articulated objects from real-world RGB images and RGB videos.

{
\vspace{1em}
\noindent \textbf{Acknowledgement:}
This work was funded in part by a Canada CIFAR AI Chair, a Canada Research Chair and NSERC Discovery Grant, and enabled in part by support from \href{https://www.westgrid.ca/}{WestGrid} and \href{https://www.computecanada.ca/}{Compute Canada}.
We thank Sanjay Haresh for help with scanning and narration for our video, Yue Ruan for scanning and data annotation, and Supriya Pandhre, Xiaohao Sun, and Qirui Wu for help with data annotation.
}

{\small
\bibliographystyle{plainnat}
\setlength{\bibsep}{0pt}
\bibliography{main}
}

\newpage
\appendix

In this supplement to the main paper we provide implementation details (\Cref{sec:supp:implementation}), additional quantitative and qualitative results (\Cref{sec:supp:results}), as well as details on the dataset statistics and construction (\Cref{sec:supp:data}).

\section{Additional implementation details}
\label{sec:supp:implementation}

\subsection{Training loss details}
\label{sec:supp:loss}

\mypara{Detection and Segmentation Losses.}
Our architecture uses the Mask R-CNN~\cite{he2017mask} loss which is composed of losses for the RPN module, part detection and segmentation losses for each proposed region of interest: 
$L_\text{Mask R-CNN} = L_\text{rpn} + L_\text{det} + L_\text{seg}$.
We refer the reader to the original paper for details on the implementation of these losses.

\mypara{Motion Losses.}
We extend the Mask R-CNN network with extra heads for motion prediction.  To train the motion prediction heads, we add additional loss terms to the loss associated with each ROI.
We construct the motion loss $L_m$ as a weighted sum of cross-entropy loss for the motion type ($L_c$), and smooth L1 losses for regressing the motion axis ($L_a$) and motion origin ($L_o$): $L_m = \lambda_c L_c + \lambda_a L_a + \lambda_o L_o$.  The motion loss terms for each RoI $i$ are given by:
\begin{equation}
    \begin{split}
        L_{c_i} &= \lossce(\hat{c}_i, c_i) \\
        L_{a_i} &= \losshuber(\hat{a}_i, a_i) \\
        L_{o_i} &= \losshuber(\hat{o}_i, o_i) \indicator\{c_i = \text{rotation}\}
    \end{split}
    \label{eq:cc_additional_loss}
\end{equation}
where $\hat{c}_i$ is the predicted motion type and $c_i$ is the ground truth motion type, $\hat{a}_i$ is the predicted axis and $a_i$ is the ground truth axis, $\hat{o}_i \in \R^3$ is the predicted origin and $o_i$ is the ground truth origin.  
We set $\lambda_c = 1, \lambda_a = 8, \lambda_o = 8$ for our experiments.
For \textbf{\opdnetcc}, the motion axis and origin are in camera coordinates.
The overall loss is given by $L_{\text{CC}} = L_\text{Mask R-CNN} + L_m$.   
\textbf{\opdnetocs} has the same additional loss as \opdnetcc for motion parameters, but with additional smooth L1 loss for the extrinsic matrix $L_{\text{ext}}$.  
\begin{equation}
    L_{\text{OC-S}} = L_{\text{CC}} + L_{\text{ext}_s} = L_{\text{CC}} + \lambda_\text{ext} \losshuber(\hat{e}_s, e_s) 
    \label{eq:ocs_total_loss}
\end{equation}
We represent the extrinsic matrix as a vector $e_s$ of length 12 (9 for rotation, 3 for translation). The extrinsic matrix $e_s$ is predicted by taking the features for the entire images directly from the backbone network.  
For \opdnetocs, the motion axis and motion origin are in the canonical object coordinate instead of the camera coordinate.  We set $\lambda_\text{ext} = 15$ for our experiments.

\subsection{Coordinate system details}

We use three coordinate systems in our experiments: i) camera coordinates; ii) canonical object coordinates (equivalent to world coordinates for synthetic data); and iii) ANOCs (anistropically-scaled normalized object coordinates), our adaptation of the normalized object coordinates (NOCs).

\mypara{Camera Coordinates.}
Our task is to predict the motion parameters from a single-view image so camera coordinates are a natural coordinate system.
We evaluate all motion parameters in camera coordinates.
The input point clouds for \ancsh and our \pnetopd baseline are also represented in camera coordinates.

\mypara{Canonical Object Coordinates.}
Inspired by the canonical coordinates used in the \ancsh~\cite{li2020category} approach. 
We use canonical object coordinate in our \opdnetocs model to predict the motion axis and motion origin in a more consistent frame of reference.
To obtain a canonical object coordinate frame we either rely on existing alignments of objects to a canonical pose (for \ourdatacad), or annotate a semantically-consistent oriented bounding box (OBB) with a consistent front and up axis for each object (for \ourdatareal).

\mypara{ANOCs (anistropically-scaled NOCs).}
The ANOCs coordinate system further normalizes the canonical object coordinates.
We use the dimensions of the bounding box of each object to normalize each dimension to $[-0.5,0.5]$.
This makes it easier to define the candidate motion origins for the \randmot and \mostfreq baselines.

\begin{figure*}[ht]
\centering
\includegraphics[width=0.8\linewidth]{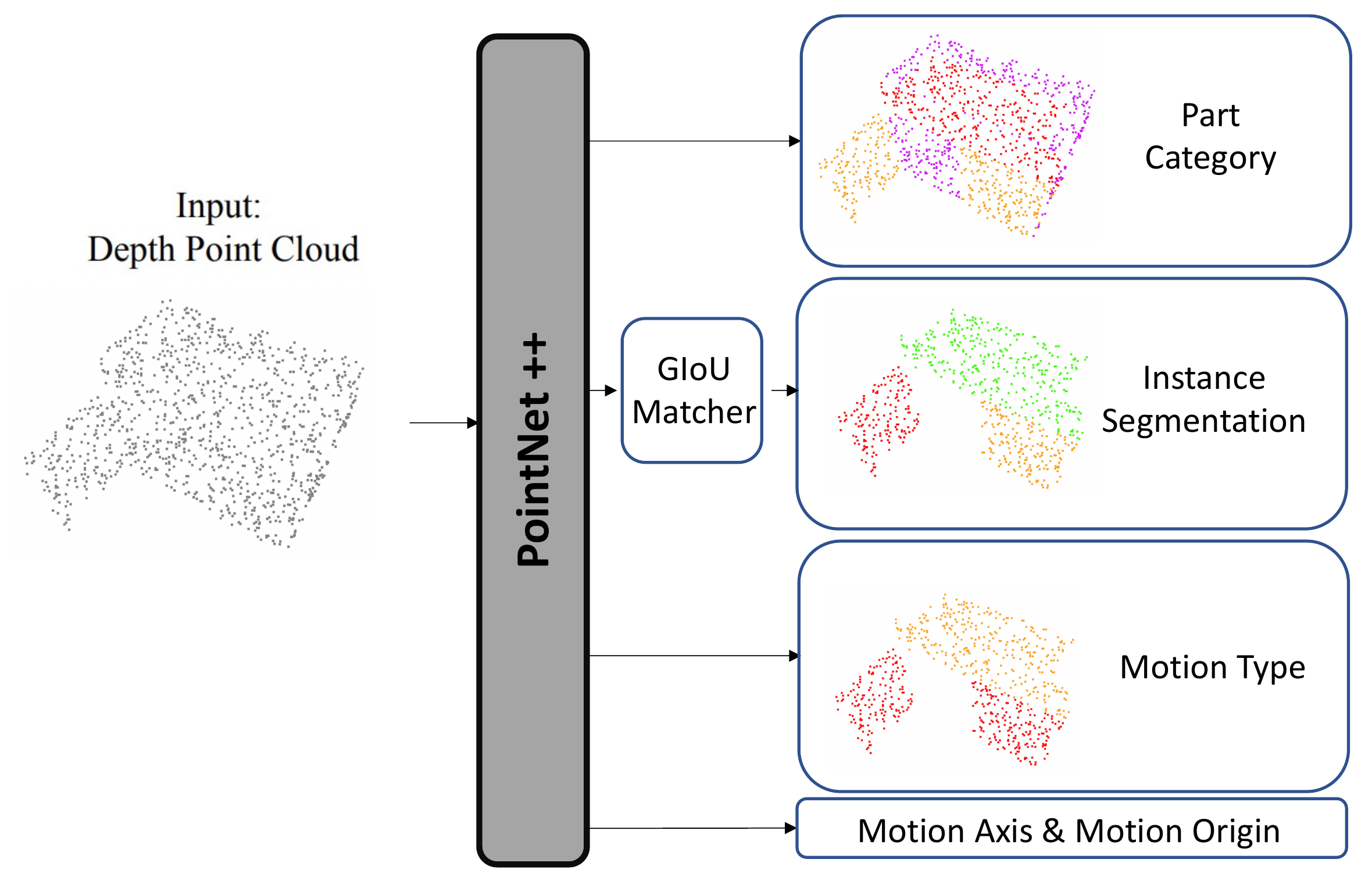}
\caption{Network structure for \pnetopd. Given the single-view point cloud, the network predicts part category, instance segmentation, motion type, motion axis and motion origin. For the part category, the base part of the object is one category. For other predictions we only consider predicting the moving parts. For the motion axis and motion origin, we assign the motion to each moving part, and all points in that moving part have the same ground truth motion type, motion axis and motion origin. We use GIoU~\cite{rezatofighi2018giou} and the Hungarian algorithm to match part instances from the predictions with the ground truth for the evaluation.}
\label{fig:network_pc}
\end{figure*}

\subsection{\pnetopd baseline architecture}

The \pnetopd baseline uses a PointNet++~\cite{qi2017pointnetplusplus} architecture to process a single-view 3D point cloud of the object.
We use a set of part category labels corresponding to the openable part types with one additional label representing any other parts that are not articulated.
We also predict an instance segmentation id to separate part instances.
These ids do not have a natural ordering defining correspondences between parts in different objects in the dataset, so we instead match the predicted instance id with instances in the ground truth using GIoU~\cite{rezatofighi2018giou} and the Hungarian algorithm.
We use a mIoU loss for the part category, instance id and motion type.
For the motion axis and motion origin we use an MSE loss.

\subsection{Re-implementation of ANCSH}

\begin{table*}[t]
\caption{Comparison between our re-implementation of ANCSH~\cite{li2020category} and the results reported in the original paper on the original eyeglasses dataset.}
\centering{
\begin{tabular}{@{} l rrrrr @{}}
\toprule
& \multicolumn{3}{c}{Part-based Metrics} & \multicolumn{2}{c}{Joint Parameters} \\
\cmidrule(lr){2-4} \cmidrule(lr){5-6} 
Method & Rotation Err $\downarrow$ & Translation Err $\downarrow$ & 3D IoU $\% \uparrow$ & Angle Err $\downarrow$ & Distance Err $\downarrow$ \\
\midrule
\citet{li2020category} & 3.7, 5.1, 3.7 & \textbf{0.035}, \textbf{0.051}, \textbf{0.057} & \textbf{87.4}, 43.6, 44.5 & 2.2, \textbf{2.3} & \textbf{0.019}, \textbf{0.014} \\
Our implementation & \textbf{2.8}, \textbf{2.8}, \textbf{3.5} & 0.039, 0.053, 0.072 & 87.0, \textbf{45.6}, \textbf{45.5} & \textbf{2.1}, 2.5 & 0.023, 0.024 \\
\bottomrule
\end{tabular}
}
\label{tab:results-ancsh-reimplementation}
\end{table*}

We re-implemented the ANCSH approach by \citet{li2020category} in PyTorch.
\Cref{tab:results-ancsh-reimplementation} reports the results of our re-implementation against the original reported results.
We see that our re-implementation gives comparable results with the original, with small variations (performance under some metrics improved while it is slightly worse along some other metrics).
We performed this comparison as a sanity check experiment to confirm that our re-implementation is consistent with the results reported by the authors.

\section{Additional results}
\label{sec:supp:results}

\begin{table}
\caption{
Part detection and segmentation results on the val set. $\apbb$ is the mAP for the 2D bounding box, and $\ap$ is the mAP for the instance segmentation.
}
\centering
\resizebox{\linewidth}{!}{
\begin{tabular}{@{}ll rrr rrr@{}}
\toprule
& & \multicolumn{3}{c}{Detection} & \multicolumn{3}{c}{Segmentation} \\
\cmidrule(lr){3-5} \cmidrule(lr){6-8}
Input & Model & \apbb & $\apbb_{50}$ & $\apbb_{75}$ & \ap & $\ap_{50}$ & $\ap_{75}$  \\
\midrule
\multirow{2}{*}{RGB}
& \opdnetcc & 50.5 & 74.7 & 55.5 & 45.1 & 67.1 & 50.2 \\
& \opdnetocs & \textbf{50.6} & \textbf{75.3} & \textbf{56.1} & \textbf{45.5} & \textbf{67.9} & \textbf{50.6}\\
\midrule
\multirow{2}{*}{D}
& \opdnetcc & \textbf{44.5} & 69.4 & \textbf{48.2} & \textbf{38.8} & 60.6 & \textbf{42.0}\\
& \opdnetocs & 44.1 & \textbf{70.5} & 46.8 & 38.3 & \textbf{61.3} & 40.9\\
\midrule
\multirow{2}{*}{RGBD}
& \opdnetcc & \textbf{48.6} & \textbf{73.6} & \textbf{52.5} & \textbf{42.3} & \textbf{65.3} & \textbf{45.7}\\
& \opdnetocs & 47.1 & 73.3 & 50.9 & 41.1 & 64.0 & 44.6\\
\bottomrule
\end{tabular}
}
\label{tab:results-bbseg}
\end{table}

\subsection{Part detection and segmentation performance}
\label{sec:supp:results-part-seg}

In \Cref{tab:results-bbseg}, we report the standard COCO metrics for segmentation including \ap (averaged over IoU $0.50$ to $0.95$ thresholds at a increment of $0.5$), \ap$_{50}$ (for IoU=0.5), \ap$_{75}$ (for IoU$=0.75$).  We also report the mean average precision for part detection over the 2D bounding boxes (\apbb, \apbb$_{50}$, \apbb$_{70}$).
All results are on the validation set for the \textit{pretrained} part detector.

\begin{table*}
\caption{Comparison against baselines on the \ourdatacad test set with metrics broken down by motion type.
}
\resizebox{\linewidth}{!}{
\begin{tabular}{@{}ll rrrrr r rr@{}}
\toprule
& & \multicolumn{3}{c}{Part-averaged mAP $\% \uparrow$} & \multicolumn{5}{c}{Motion-averaged mAP $\% \uparrow$} \\
\cmidrule(lr){3-5} \cmidrule(lr){6-10} 
Input & Model & \partdet & +\mtype & +\mtype{}\axis{}\orig & \motiondet & +\mtype{}\axis & +\mtype{}\axis (t) & +\mtype{}\axis (r) & +\mtype{}\orig (r)   \\
\midrule
\multirow{4}{*}{RGBD}
& \randmot & 5.1 & 1.4 & 0.2 & 6.5 & 0.7 & 1.0 & 0.9 & 0.7 \\
& \mostfreq & 69.4 & 66.1 & 27.8 & 73.6 & 61.6 & 61.3 & 62.4 & 22.1 \\
\cmidrule(lr){2-10}
& \opdnetcc & \textbf{69.6} & 67.4 & 41.6 & \textbf{75.4} & 55.9 & 57.6 & 54.9 & \textbf{74.6} \\
& \opdnetocs & 69.4 & \textbf{67.9} & \textbf{49.7} & 75.3 & \textbf{66.3} & \textbf{65.0} & \textbf{68.6} & 73.4 \\
\midrule
\multirow{4}{*}{D}
& \randmot & 4.8 & 1.2 & 0.1 & 5.7 & 0.6 & 0.9 & 0.8 & 0.8 \\
& \mostfreq & 66.7 & 64.1 & 25.5 & 69.9 & 58.5 & 58.5 & 59.8 & 19.4 \\
\cmidrule(lr){2-10}
& \opdnetcc & \textbf{67.9} & \textbf{66.6} & 38.6 & \textbf{72.9} & 52.7 & 54.2 & 52.3 & \textbf{71.9} \\
& \opdnetocs & 66.7 & 65.0 & \textbf{47.3} & 71.9 & \textbf{62.3} & \textbf{60.9} & \textbf{65.3} & 70.7 \\
\midrule
\multirow{4}{*}{RGB}
& \randmot & 4.7 & 1.2 & 0.1 & 5.6 & 0.6 & 0.8 & 0.8 & 0.6 \\
& \mostfreq & 66.0 & 63.5 & 27.3 & 71.8 & 60.5 & 61.2 & 60.3 & 20.9 \\
\cmidrule(lr){2-10}
& \opdnetcc & \textbf{67.2} & \textbf{66.0} & 38.3 & \textbf{75.3} & 53.5 & 56.9 & 50.7 & 70.7 \\
& \opdnetocs & 66.0 & 64.8 & \textbf{46.8} & 73.9 & \textbf{63.6} & \textbf{65.0} & \textbf{63.2} & \textbf{71.1} \\
\bottomrule
\end{tabular}
}
\label{tab:results_test}
\end{table*}

\subsection{Motion parameter performance by motion type}

In \Cref{tab:results_test}, we present the results using our best model on the \ourdatacad test set.
We include a breakdown of the motion parameter estimation by motion type for translation (t) and rotation (r).
Note that the motion origin is only valid for rotation.
Our \opdnetocs outperforms \opdnetcc in all cases for motion axis prediction but slightly underperforms for origin prediction.

\begin{table*}
\caption{
Error metrics for matched instances for \ourdatacad test set (micro-averaged)
with the predicted part matched to the ground-truth at IoU of 0.5 and matching motion type. 
}
\resizebox{\linewidth}{!}{
\begin{tabular}{@{} ll rrrr rrr @{}}
\toprule
&  & \multicolumn{4}{c}{Error $\downarrow$} & \multicolumn{3}{c}{\#Matched $\uparrow$} \\
 \cmidrule(l{2pt}r{0pt}){3-6} \cmidrule(l{2pt}r{0pt}){7-9} 
Input & Model & \axis & \axis(t) & \axis(r) & \orig & \axis & \axis(t) & \axis(r)/\orig \\
\midrule
\multirow{5}{*}{RGBD}
& \randmot & 59.71 & 59.06 & 60.21 & 0.38 & 44713 & 19598 & 25115 \\
& \mostfreq & 11.08 & 3.57 & 16.33 & 0.32 & 44896 & 18468 & 26428 \\
\cmidrule(l{0pt}r{0pt}){2-9}
& \opdnetcc & 9.4$\pm$0.02 &	6.7$\pm$0.13 &	11.5$\pm$0.09 &	0.1$\pm$0 &	46019.3$\pm$59.73 &	20123.8$\pm$11.93 &	25895.5$\pm$68.12 \\
& \opdnetocs & 6.9$\pm$0.07 &	4.1$\pm$0.08 &	\markbest{9.0}$\pm$0.1 &	0.1$\pm$0 &	46250.4$\pm$58.47 &	20133.4$\pm$82.60 &	26117$\pm$88.61 \\
\midrule
\multirow{2}{*}{D (PC)}
& \ancsh~\cite{li2020category} & 10.41 & - & 10.41 & 0.09 & 6935 & - & 6935 \\
& \pnetopd & \markbest{6.59} & \markbest{3.38} & 9.11 & 0.09 & 19672 & 8666 & 11006 \\
\cmidrule(l{0pt}r{0pt}){2-9}
\multirow{3}{*}{D}
& \opdnetcc & 9.6$\pm$0.08 &	6.3$\pm$0.09 &	12$\pm$0.11 &	0.1$\pm$0 &	45055$\pm$60.28 &	19249.6$\pm$52.59 &	25805.4$\pm$57.31  \\
& \opdnetocs & 7.1$\pm$0.10 &	4.1$\pm$0.10 &	9.3$\pm$0.14 &	0.1$\pm$0 &	45537.2$\pm$57.13 &	19494$\pm$102.17 &	26043.2$\pm$73.82 \\
\midrule
\multirow{3}{*}{RGB}
& \opdnetcc & 9.7$\pm$0.05 &	6.7$\pm$0.07 &	12.1$\pm$0.09 &	0.1$\pm$0 &	46282$\pm$92.65 &	20424$\pm$81.72 &	25858$\pm$22.91 \\
& \opdnetocs & 7.4$\pm$0.09 &	4.1$\pm$0.04 &	9.9$\pm$0.17 &	0.1$\pm$0 &	46545$\pm$128.38 &	20486$\pm$73.61 &	26059$\pm$81.78 \\
\bottomrule
\end{tabular}
}
\label{tab:results-cad-test-error-micro}
\end{table*}

\begin{table*}
\caption{Error metrics for matched instances for \ourdatacad test set (motion averaged) with matches determined by sweeping over different IoU thresholds.}
\resizebox{\linewidth}{!}{
\begin{tabular}{@{} ll rrrr rrr@{}}
\toprule
& & \multicolumn{7}{c}{Motion-averaged} \\
\cmidrule(l{2pt}r{0pt}){3-9}
& & \multicolumn{4}{c}{Error $\downarrow$} & \multicolumn{3}{c}{\#Matched $\uparrow$} \\
\cmidrule(l{0pt}r{2pt}){3-6} \cmidrule(l{2pt}r{0pt}){7-9} 
Input & Model & \axis & \axis(t) & \axis(r) & \orig & \axis & \axis(t) & \axis(r)/\orig \\
\midrule
\multirow{5}{*}{RGBD}
& \randmot & 59.63 & 59.06 & 60.2 & 0.38 & 22493 & 19853 & 12566 \\
& \mostfreq & 9.96 & 3.57 & 16.36 & 0.32 & 22505 & 18563 & 13223 \\
\cmidrule(l{0pt}r{0pt}){2-9}
& \opdnetcc & 	9.1$\pm$0.03 &	6.7$\pm$0.13 &	11.5$\pm$0.09 &	0.1$\pm$0.00 &	23113.3$\pm$27.6 &	20307.3$\pm$15.38 &	12959.8$\pm$34.08 \\
& \opdnetocs & 6.6$\pm$0.06 &	4.1$\pm$0.08 &	\markbest{9.0}$\pm$0.10 &	0.1$\pm$0.00 &	23239.2$\pm$31.33 &	20335.2$\pm$81.92 &	13071.8$\pm$44.58 \\
\midrule
\multirow{2}{*}{D (PC)}
& \ancsh~\cite{li2020category} & 10.36 & - & 10.36 & 0.09 & 6975 & - & 6975  \\
& \pnetopd & \markbest{6.25} & \markbest{3.38} & 9.12 & 0.09 & 9862 & 8705 & 5510 \\
\cmidrule(l{0pt}r{0pt}){2-9}
\multirow{3}{*}{D}
& \opdnetcc & 9.2$\pm$0.08 &	6.3$\pm$0.09 &	12.0$\pm$0.11 &	0.1$\pm$0.00 &	22625.4$\pm$23.91 &	19432.0$\pm$51.54 &	12909.2$\pm$28.68 \\
& \opdnetocs & 6.7$\pm$0.10 &	4.1$\pm$0.10 &	9.3$\pm$0.14 &	0.1$\pm$0.00 &	22888.4$\pm$30.77 &	19713$\pm$105.68 &	13031.8$\pm$36.28 \\
\midrule
\multirow{3}{*}{RGB}
& \opdnetcc & 9.4$\pm$0.05 &	6.7$\pm$0.07 &	12.1$\pm$0.10 &	0.1$\pm$0.00 &	23256.8$\pm$44.0 &	20638.6$\pm$75.84 &	12937.6$\pm$11.57 \\
& \opdnetocs & 7.0$\pm$0.08 &	4.1$\pm$0.04 &	9.9$\pm$0.17 &	0.1$\pm$0.00 &	23398.6$\pm$68.99 &	20718.2$\pm$81.51 &	13039.4$\pm$41.66 \\
\bottomrule
\end{tabular}
}
\label{tab:results-cad-test-error-motion-ave}
\end{table*}

\begin{table}
\caption{
Error metrics for the \ourdatareal test set (micro-averaged)
with the predicted part matched to the ground truth at IoU of 0.5 and matching motion type.
}
\centering
\scriptsize{
\resizebox{\linewidth}{!}{
\begin{tabular}{@{} ll rrrr rrr@{}}
\toprule
&  & \multicolumn{4}{c}{Error $\downarrow$} & \multicolumn{3}{c}{\#Matched $\uparrow$}  \\
\cmidrule(l{0pt}r{2pt}){3-6} \cmidrule(l{2pt}r{0pt}){7-9} 
Input & Model & \axis & \axis(t) & \axis(r) & \orig & \axis & \axis(t) & \axis(r)/\orig \\
\midrule
\multirow{4}{*}{RGBD}
& \randmot & 59.82 & 59.62 & 60.12 & 0.38 & 10088 & 5990 & 4098 \\
& \mostfreq & 13.99 & \markbest{7.67} & 22.40 & 0.30 & 9738 & 5562 & 4176 \\
\cmidrule(l{0pt}r{0pt}){2-9}
& \opdnetcc & 15.21 & 16.07 & 13.84 & 0.10 & 9942 & 6119 & 3823 \\
& \opdnetocs & 9.84 & 8.83 & 11.44 & 0.14 & 10076 & 6191 & 3885 \\
\midrule
\multirow{3}{*}{D}
& \pnetopd & \markbest{7.33} & \markbest{7.67} & \markbest{6.99} & 0.08 & 4461 & 2265 & 2196 \\
& \opdnetcc & 22.37 & 26.70 & 15.56 & 0.12 & 9485 & 5800 & 3685 \\
& \opdnetocs & 13.76 & 13.67 & 13.89 & 0.17 & 9417 & 5738 & 3679 \\
\midrule
\multirow{3}{*}{RGB}
& \opdnetcc & 14.93 & 15.60 & 13.84 & 0.12 & 9916 & 6151 & 3765 \\
& \opdnetocs & 10.32 & 9.32 & 11.91 & 0.16 & 10225 & 6270 & 3955 \\
\bottomrule
\end{tabular}
}
}
\label{tab:results-real-test-error-micro}
\end{table}

\begin{table}
\caption{Error metrics for matched instances for \ourdatareal test set (motion averaged) with matches determined by sweeping over different IoU thresholds.}
\centering
\scriptsize{
\resizebox{\linewidth}{!}{
\begin{tabular}{@{} ll rrrr rrr@{}}
\toprule
& & \multicolumn{7}{c}{Motion-averaged} \\
\cmidrule(l{0pt}r{2pt}){3-9}
& & \multicolumn{4}{c}{Error $\downarrow$} & \multicolumn{3}{c}{\#Matched $\uparrow$} \\
\cmidrule(l{0pt}r{2pt}){3-6} \cmidrule(l{2pt}r{0pt}){7-9} 
Input & Model & \axis & \axis(t) & \axis(r) & \orig & \axis & \axis(t) & \axis(r)/\orig \\
\midrule
\multirow{4}{*}{RGBD}
& \randmot & 59.88 & 59.65 & 60.12 & 0.38 & 5074 & 6050 & 2049   \\
& \mostfreq & 15.03 & \markbest{7.65} & 22.40 & 0.30 & 4887 & 5597 & 2088    \\
\cmidrule(l{0pt}r{0pt}){2-9}
& \opdnetcc & 14.97 & 16.11 & 13.84 & 0.10 & 5003 & 6184 & 1911    \\
& \opdnetocs &  10.14 & 8.84 & 11.44 & 0.14 & 5077 & 6268 & 1943    \\
\midrule
\multirow{3}{*}{D}
& \pnetopd & \markbest{7.33} & \markbest{7.67} & \markbest{6.99} & \markbest{0.08} & 2233 & 2268 & 1099    \\
& \opdnetcc & 21.14 & 26.72 & 15.56 & 0.12 & 4785 & 5881 & 1844    \\
& \opdnetocs & 13.81 & 13.71 & 13.92 & 0.17 & 4758 & 5833 & 1842    \\
\midrule
\multirow{3}{*}{RGB}
& \opdnetcc & 14.72 & 15.61 & 13.84 & 0.12 & 4984 & 6202 & 1883    \\
& \opdnetocs & 10.60 & 9.29 & 11.91 & 0.16 & 5146 & 6337 & 1978    \\
\bottomrule
\end{tabular}
}
}
\label{tab:results-real-test-error-motion-ave}
\end{table}

\subsection{Part motion estimation error metrics}

We also evaluate motion parameter estimation by computing the angle error for axis predictions, and normalized distance error for origin predictions following prior work~\cite{li2020category,wang2019shape2motion}.
As we noted in the main paper (Section 3.2), these error metrics are only computed for matched parts, and do not consider that the number of matched parts may differ between models (e.g., if a model detects only one part the error metric will only take that part into account).
For models that predict both parts and their motion parameters, predicting more parts may be penalized for attempting to predict motion parameters of challenging parts.

We consider two ways of computing the error metrics: 1) we compute the micro-averaged mean of errors for detected parts (with \texttt{maxDet=100}) that are matched to ground truth parts at IoU of 0.5); and 2) we compute the average error across different IoU thresholds, different area and different \texttt{maxDet} to determine the matching between the prediction and GT. In this setting, we compute the average error for each motion type and then compute the macro-average (across motion types) to obtain the final average error.
We report the error as well as the average number of matched parts for the two settings for both the \ourdatacad (\Cref{tab:results-cad-test-error-micro,tab:results-cad-test-error-motion-ave}), and \ourdatareal(\Cref{tab:results-real-test-error-micro,tab:results-real-test-error-motion-ave}).
We find that there is no noticeable difference between the two settings, and that the micro-averaged error at IoU=0.5 is reflective of the overall error that sweeps across multiple IoUs.

From \Cref{tab:results-cad-test-error-micro,tab:results-real-test-error-micro}, we see that \opdnetcc and \opdnetoc have comparable number of matched parts with \opdnetoc having lower motion parameter errors (the trend holds across the different inputs).
In contrast, \pnetopd has the lowest axis error but also has fewer matched parts.
Because we trained \ancsh on only one structure (`one-door'), \ancsh has the lowest number of matched parts.
It also does not predict any motion parameters for the translation motion type.
In the next section, we examine in more detail the \ancsh performance on only the `one-door' kinematic structure.

\subsection{Comparison against ANCSH}

\begin{table*}
\caption{
Comparison of \opdnetocs against \ancsh\cite{li2020category} and \pnetopd baselines.
`Complete set' means evaluation on all objects in the test set.
The `one-door set' includes only objects with one door exhibiting rotational motion.
\pnetopd is trained on objects with no more than 5 parts in our train set.
\ancsh is trained on objects in the train set with one door exhibiting rotational motion.
The `one rotating door' objects are the most frequent structure in our \ourdatacad dataset.
\opdnetocs is trained with all objects in the train set for the `complete' test, and trained on only single rotating door objects.}
\resizebox{\linewidth}{!}{
\begin{tabular}{@{} ll rrrr rrr  |rrrr rrr@{}}
\toprule
& & \multicolumn{4}{c}{Part-averaged} & \multicolumn{10}{c}{Motion-averaged} \\
\cmidrule(l{0pt}r{2pt}){3-6} \cmidrule(l{2pt}r{0pt}){7-16}
& & \multicolumn{4}{c}{mAP $\% \uparrow$} & \multicolumn{3}{c}{mAP $\% \uparrow$} & \multicolumn{4}{c}{Error $\downarrow$} & \multicolumn{3}{c}{\#Matched $\uparrow$} \\
\cmidrule(l{0pt}r{2pt}){3-6} \cmidrule(l{2pt}r{0pt}){7-9} \cmidrule(l{2pt}r{0pt}){10-13} \cmidrule(l{2pt}r{0pt}){14-16} 
Input & Model & \partdet & +\mtype & +\mtype{}\axis & +\mtype{}\axis{}\orig & \motiondet & +\mtype{}\axis & +\mtype{}\axis{}\orig & \axis & \axis(t) & \axis(r) & \orig & \axis & \axis(t) & \axis(r)/\orig \\
\midrule
Complete Set \\
\midrule
RGBD & \randmot & 5.0 & 1.3 & 0.2 & 0.1 & 6.2 & 0.7 & 0.3     & 59.63 & 59.06 & 60.2 & 0.38 & 22493 & 19853 & 12566    \\
RGBD & \mostfreq & \textbf{69.4} & 66.1 & 49.2 & 27.8 & 73.6 & 61.6 & 38.8     & 9.96 & 3.57 & 16.36 & 0.32 & 22505 & 18563 & 13223    \\
D (PC)& \pnetopd & 20.4 & 19.3 & 14.0 & 13.6 & 22.0 & 18.1 & 17.6     & \textbf{6.25} & \textbf{3.38} & 9.12 & 0.09 & 9862 & 8705 & 5510    \\
D (PC)& \ancsh & 2.7 & 2.7 & 2.3 & 2.1 & 3.9 & 3.1 & 2.8     & 10.36 & - & 10.36 & 0.09 & 6975 & - & 6975    \\ 
RGB & \opdnetocs & 67.5 &	66.4 &	51.5 &	47.9 &	75.1 &	64.6 &	62.2 &	6.96 &	4.2 &	9.72 &	0.11 &	23401 &	20682 &	13060 \\
D & \opdnetocs & 67.6 &	65.8 &	52.5 &	48.6 &	72.2 &	63.2 &	60.8 &	6.54 &	4.1 &	\textbf{8.98} &	0.11 &	22945 &	19910 &	12990 \\
RGBD & \opdnetocs & \textbf{69.4} &	\textbf{67.9} &	\textbf{53.5} &	\textbf{49.7} &	\textbf{75.3} &	\textbf{66.3} &	\textbf{63.7} &	6.47 &	3.91 &	9.02 &	0.10 &	23247 &	20288 &	13103 \\
\midrule
One-Door Set \\
\midrule
RGBD & \randmot & 14.6 & 4.3 & 1.3 & 0.0 & 3.6 & 0.8 & 0.0     & 60.58 & - & 60.58 & 0.36 & 3224 & - & 3224    \\
RGBD & \mostfreq & \textbf{96.3} & \textbf{96.3} & 41.4 & 6.6 & \textbf{96.3} & 41.4 & 6.6     & 30.17 & - & 30.17 & 0.32 & 4775 & - & 4775    \\
D (PC)& \pnetopd & 71.1 & 71.1 & 45.6 & 40.0 & 71.1 & 45.6 & 40.0     & 11.55 & - & 11.55 & 0.11 & 4075 & - & 4075    \\
D (PC)& \ancsh & 84.2 & 84.2 & \textbf{75.2} & \textbf{70.0} & 84.2 & \textbf{75.2} & \textbf{70.0}     & \textbf{5.92} & - & \textbf{5.92} & \textbf{0.06} & 4465 & - & 4465    \\
RGB & \opdnetocs & 90.0 & 90.0 & 65.9 & 54.8 & 90.0 & 65.9 & 54.8     & 11.31 & - & 11.31 & 0.14 & 4582 & - & 4582    \\
D & \opdnetocs & 94.8 & 94.8 & 69.2 & 61.2 & 94.8 & 69.2 & 61.2     & 12.77 & - & 12.77 & 0.14 & 4715 & - & 4715    \\
RGBD & \opdnetocs & \textbf{96.3} & \textbf{96.3} & 73.7 & 63.4 & \textbf{96.3} & 73.7 & 63.4     & 11.61 & - & 11.61 & 0.13 & 4775 & - & 4775    \\

\bottomrule
\end{tabular}
}
\label{tab:results-ap-motion-ave}
\end{table*}

In the main paper we evaluated the \ancsh approach of \citet{li2020category} on a dataset including objects with varying number of parts and motion types.
This puts this approach at a disadvantage as it was designed such that each trained model can only operate on a fixed kinematic chain.
Thus, to evaluate \ancsh in a setting that is closer to its assumption of a fixed kinematic chain, we construct a `one-door dataset' with includes 243 instances with one rotating door part from our original 683 models.
More specifically, we pick 172 models from train, 32 models from val and 39 models from test.
We train \ancsh, \pnetopd, and \method on the train set of the one-door dataset and evaluate on the test set of the `one-door dataset'.

\Cref{tab:results-ap-motion-ave} shows the results on the one-door dataset.
Unlike the results on the complete set, \ancsh performs well on this subset of the data because the strong assumption of a single kinematic structure is satisfied.
In addition, \ancsh is given prior knowledge about the rest (closed) state of the object during training, and requires additional annotation that our method does not need.
Notably, under this setting, \ancsh has a much more accurate rotation origin prediction than our methods.
While our methods have slightly less accurate motion parameter estimation, they have more accurate part detection and our structure-agnostic approach handles arbitrary variations in kinematic structure with the same model.

\subsection{Additional analysis}

\begin{table*}[t]
\caption{Analysis of performance of \opdnetoc given ground truth 2D bounding box, part category and object pose on the validation set of \ourdatacad.  We compare the performance of using the predicted part vs using the ground truth 2D bounding box and part category (\gtboxpart), the ground truth object pose (\gtpose), and the combination (\gtboxpartpose).  We see that having access to the ground truth object pose (\gtpose) is important for accurate motion prediction.}
\resizebox{\linewidth}{!}{
\begin{tabular}{@{} ll rrrr rrr @{}}
\toprule
& & \multicolumn{4}{c}{Part-averaged mAP $\% \uparrow$} & \multicolumn{3}{c}{Motion-averaged mAP $\% \uparrow$} \\
\cmidrule(l{0pt}r{2pt}){3-6} \cmidrule(l{2pt}r{0pt}){7-9}
 & & \partdet & +\mtype & +\mtype{}\axis & +\mtype{}\axis{}\orig & \motiondet & +\mtype{}\axis & +\mtype{}\axis{}\orig \\
\midrule
\multirow{4}{*}{RGBD}
& \opdnetoc & 72.5$\pm$0.34 & 70.6$\pm$0.29 & 51.7$\pm$0.62 & 47.1$\pm$0.59 & 75.4$\pm$0.07 & 61.6$\pm$0.32 & 59.0$\pm$0.32 \\
& \gtboxpart & \markbest{99.0}$\pm$0.00 &	\markbest{90.9}$\pm$0.16 & 50.6$\pm$0.36 &	45.4$\pm$0.27 &	\markbest{89.7}$\pm$0.15 &	58.1$\pm$0.32 &	54.7$\pm$0.28 \\
& \gtpose & 73.1$\pm$0.10 &	71.0$\pm$0.05 &	60.5$\pm$0.06 &	59.4$\pm$0.05 &	75.2$\pm$0.08 &	67.0$\pm$0.14 &	66.2$\pm$0.09 \\
& \gtboxpartpose & \markbest{99.0}$\pm$0.00 &	\markbest{90.6}$\pm$0.37 &	\markbest{65.5}$\pm$0.24 &	\markbest{63.8}$\pm$0.17 &	\markbest{89.5}$\pm$0.19 &	\markbest{73.3}$\pm$0.26 &	\markbest{72.0}$\pm$0.30 \\
\midrule
\multirow{4}{*}{D}
& \opdnetoc & 69.3$\pm$0.35 & 67.5$\pm$0.33 & 50.7$\pm$0.55 & 45.1$\pm$0.50 & 72.5$\pm$0.26 & 59.1$\pm$0.36 & 55.9$\pm$0.49 \\
& \gtboxpart & \markbest{99.0}$\pm$0.00 &	\markbest{89.7}$\pm$0.30 &	50.6$\pm$0.09 &	45.2$\pm$0.17 &	\markbest{88.9}$\pm$0.26 &	57.8$\pm$0.19 &	54.3$\pm$0.26 \\
& \gtpose & 70.1$\pm$0.21 &	68.3$\pm$0.22 &	59.0$\pm$0.14 &	57.9$\pm$0.13 &	73.3$\pm$0.11 &	65.2$\pm$0.09 &	64.4$\pm$0.10 \\
& \gtboxpartpose & \markbest{99.0}$\pm$0.00 &	\markbest{89.3}$\pm$0.23 &	\markbest{64.8}$\pm$0.19 &	\markbest{63.1}$\pm$0.09 &	\markbest{88.4}$\pm$0.33 &	\markbest{72.7}$\pm$0.29 &	\markbest{71.5}$\pm$0.29 \\
\midrule
\multirow{4}{*}{RGB}
& \opdnetoc & 74.2$\pm$0.34 & 72.4$\pm$0.32 & 52.4$\pm$0.31 & 47.3$\pm$0.40 & 79.1$\pm$0.24 & 62.6$\pm$0.40 & 59.6$\pm$0.45 \\
& \gtboxpart & \markbest{99.0}$\pm$0.00 &	\markbest{91.3}$\pm$0.14 &	51.8$\pm$0.22 &	46.7$\pm$0.23 &	\markbest{90.9}$\pm$0.11 &	60.8$\pm$0.17 &	57.0$\pm$0.25 \\
& \gtpose & 75.5$\pm$0.07 &	73.6$\pm$0.09 &	61.0$\pm$0.14 &	59.8$\pm$0.08 &	79.8$\pm$0.08 &	70.5$\pm$0.06 &	69.5$\pm$0.04 \\
& \gtboxpartpose & \markbest{99.0}$\pm$0.00 &	\markbest{91.4}$\pm$0.19 &	\markbest{64.2}$\pm$0.16 &	\markbest{62.4}$\pm$0.17 &	\markbest{90.3}$\pm$0.16 &	\markbest{73.7}$\pm$0.23 &	\markbest{72.3}$\pm$0.18 \\
\bottomrule
\end{tabular}
}
\label{tab:results-cad-val-gt-full}
\end{table*}

\subsubsection{Experiments with ground truth.}
To investigate what parts of the problem are challenging, we conduct experiments using ground truth bounding boxes and part labels (\gtboxpart), as well as ground truth object pose (\gtpose) and their combination (\gtboxpartpose).
\Cref{tab:results-cad-val-gt-full} summarizes the result of using ground-truth for \opdnetoc on \ourdatacad for RGBD, D, and RGB.

For the ground truth 2D bounding boxes, we use them as proposals to extract image features for the box head and mask head in MaskRCNN (dropping all detection and segmentation losses). 
To make sure there is no gap between our training and inference, we also finetune our final model with features extracted from the ground truth bounding boxes.
As expected, when using the ground-truth bounding boxes and ground truth part label (\gtboxpart), the part detection is close to perfect.
As noted in the main paper, having the ground truth object pose (\gtpose) is more important for motion parameter estimation as seen in the increase in +\mtype{}\axis between (\gtpose) and (\gtboxpart), with further improvement when both are provided (\gtboxpartpose).
The +\mtype{}\axis metric is about the same when using the predicted bounding box vs ground truth.

\begin{figure*}[t]
  \centering
  \includegraphics[width=\linewidth]{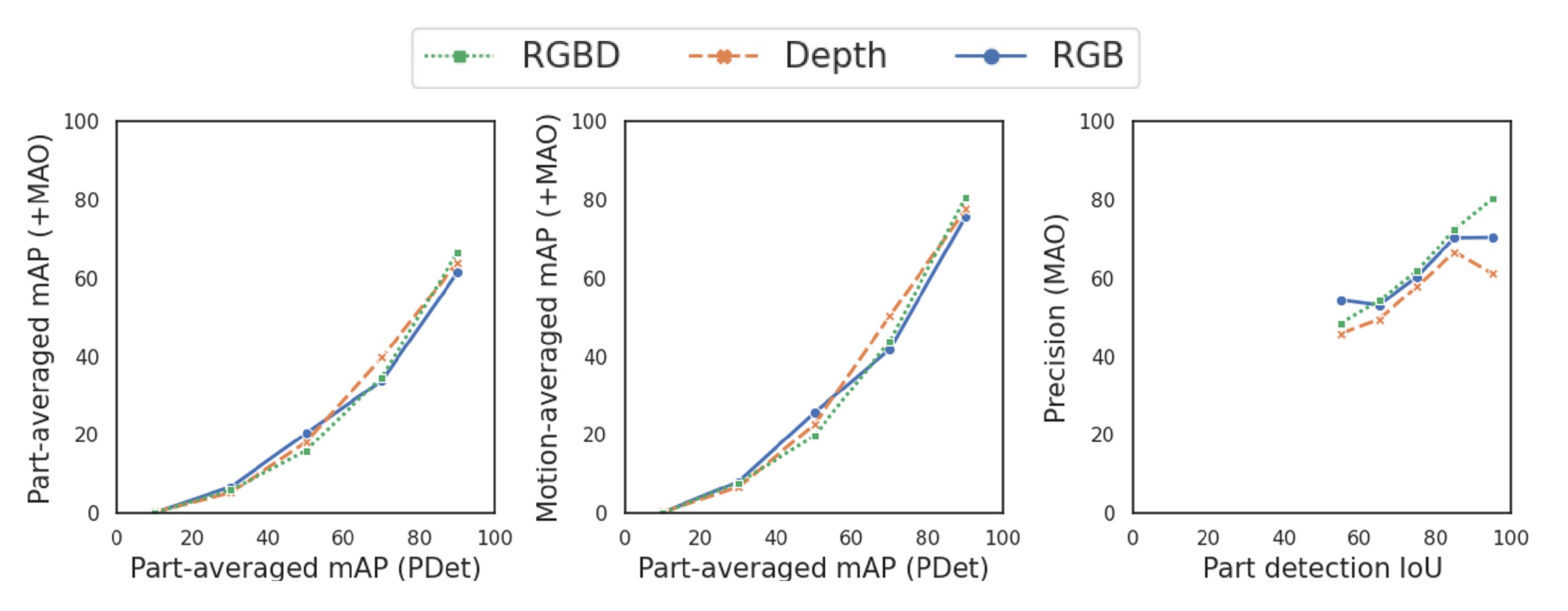}
  \caption{Plots showing correlation between part detection performance and part motion estimation performance. All results are based on the \opdnetoc model evaluated on \ourdatacad validation set. Left: part-averaged motion estimation performance against detection performance (\partdet). Middle: motion-averaged motion estimation performance against detection performance (\partdet). Right: precision of motion parameter estimation against part detection IoU. These three plots show that motion parameter estimation performance is correlated with part detection performance, with small differences between input modalities.}
  \label{fig:results_bucket}
\end{figure*}

\subsubsection{Correlation of part detection and motion estimation performance.}
We perform a more detailed analysis of the correlation between openable part detection performance and motion estimation performance.
We create plots of aggregated motion estimation performance against different buckets of part detection performance as measured by part-averaged mAP (\partdet) (see \Cref{fig:results_bucket} left and middle).
We also evaluate within the buckets divided by the IoU for each pair of GT and prediction.
\Cref{fig:results_bucket} clearly shows that for buckets with better part detection, motion parameter estimation (+\mtype{}\axis{}\orig) is also higher, which indicates that better openable part detection at the image level contributes to better motion parameter prediction. 
Although the second setting evaluates on different buckets, it is still computed at the image level, which cannot show the direct relationship between detection and motion prediction for each part.
Therefore, we design a third evaluation setting for motion prediction at the instance level.
In this setting, instead of using mAP we use precision of the motion parameter estimation on matched parts (IoU > 0.5 and part category matches).
\Cref{fig:results_bucket} (right) shows the results for the motion in different IoU buckets.
We see a strong correlation between the detection and the motion prediction.

\subsubsection{Motion thresholds.}
From the plot in \Cref{fig:results_bucket} (right) we can see that combining depth and RGB information provides a non-trivial benefit in terms of motion parameter precision when the part detection is good (RGBD results are significantly better than RGB for higher detection IoU values).
From \Cref{tab:results-motion-threshold}, we can see that when the motion threshold is stricter, the depth input and RGBD input have better motion parameter estimation results even if their detection is worse than RGB.
We hypothesize that depth information contributes to more precise motion prediction when the part detection performance is reasonable.

\begin{table*}
\caption{Results for \opdnetocs with different threshold for motion axis ($\tau_\text{axis}$) and motion origin ($\tau_\text{origin}$ times the diagonal). }
\resizebox{\linewidth}{!}{
\begin{tabular}{@{} lrr rrrr rrr rrr @{}}
\toprule
& & & \multicolumn{4}{c}{Part-averaged mAP $\% \uparrow$} & \multicolumn{6}{c}{Motion-averaged mAP $\% \uparrow$} \\
\cmidrule(l{0pt}r{2pt}){4-7} \cmidrule(l{2pt}r{0pt}){8-13}
Input & $\tau_\text{axis}$ & $\tau_\text{origin}$ & \partdet & +\mtype & +\mtype{}\axis & +\mtype{}\axis{}\orig & \motiondet & +\mtype{}\axis & +\mtype{}\axis{}\orig & +\mtype{}\axis (t) & +\mtype{}\axis (r) & +\mtype{}\orig (r) \\
\midrule
\multirow{6}{*}{RGB}
& 10 & 0.25 & 75.3 & 73.4 & 53.6 & 48.8 & 80.0 & 64.0 & 61.2 &	70.0 &	58.5 &	66.4  \\
& 10 & 0.10 & 75.3 & 73.4 & 53.6 & 35.9 & 80.0 & 64.0 & 48.8 & 70.0 & 58.5 & 32.2  \\
& 5 & 0.25  & 75.3 & 73.4 & 34.5 & 32.6 & 80.0 & 43.6 & 42.3 & 47.1 & 40.4 & 66.4 \\
& 5 & 0.10 & 75.3 & 73.4 & 34.5 & 25.2 & 80.0 & 43.6 & 34.3 & 47.1 & 40.4 & 32.2 \\
& 1 & 0.25 & 75.3 & 73.4 & 0.9 & 0.9 & 80.0 & 1.2 & 1.1 & 0.7 & 1.6 & 66.4 \\
& 1 & 0.10 & 75.3 & 73.4 & 0.9 & 0.7 & 80.0 & 1.2 & 0.9 & 0.7 & 1.6 & 32.2 \\
\midrule

\multirow{6}{*}{D}
& 10 & 0.25 & 70.5 & 68.7 & 51.7 & 46.4 & 73.4 & 60.3 & 57.6 & 63.1 &	58.9 &	64.6  \\
& 10 & 0.10 & 70.5 & 68.7 & 51.7 & 32.0 & 73.4 & 60.3 & 43.8 & 63.1 & 58.9 & 29.1 \\
& 5 & 0.25 & 70.5 & 68.7 & 35.9 & 33.3 & 73.4 & 44.8 & 43.2 & 48.6 & 41.9 & 64.6 \\
& 5 & 0.10 & 70.5 & 68.7 & 35.9 & 24.4 & 73.4 & 44.8 & 33.8 & 48.6 & 41.9 & 29.1 \\
& 1 & 0.25 & 70.5 & 68.7 & 0.9 & 0.8 & 73.4 & 1.1 & 1.1 & 1.4 & 0.9 & 64.6 \\
& 1 & 0.10 & 70.5 & 68.7 & 0.9 & 0.7 & 73.4 & 1.1 & 1.0 & 1.4 & 0.9 & 29.1 \\
\midrule

\multirow{6}{*}{RGBD}
& 10 & 0.25 & 73.3 & 71.0 & 53.6 & 48.5 & 75.4 & 62.8 & 60.0 & 64.7 &	61.7 &	67.9 \\
& 10 & 0.10  & 73.3 & 71.0 & 53.6 & 35.4 & 75.4 & 62.8 & 47.8 & 64.7 & 61.7 & 35.8 \\
& 5 & 0.25 & 73.3 & 71.0 & 37.1 & 34.6 & 75.4 & 46.2 & 44.7 & 48.1 & 44.9 & 67.9 \\
& 5 & 0.10 & 73.3 & 71.0 & 37.1 & 26.8 & 75.4 & 46.2 & 36.5 & 48.1 & 44.9 & 35.8 \\
& 1 & 0.25 & 73.3 & 71.0 & 1.5 & 1.5 & 75.4 & 2.0 & 2.0 & 1.8 & 2.2 & 67.9 \\
& 1 & 0.10 & 73.3 & 71.0 & 1.5 & 1.2 & 75.4 & 2.0 & 1.7 & 1.8 & 2.2 & 35.8 \\
\bottomrule
\end{tabular}
}
\label{tab:results-motion-threshold}
\end{table*}

\subsection{Additional qualitative results}

\label{sec:supp:qualitative}

\begin{figure*}
\centering
\setkeys{Gin}{width=\linewidth}
\begin{tabularx}{\textwidth}{Y Y Y Y Y Y Y Y Y}
\toprule

\multicolumn{9}{c}{\ourdatacad}\\
\midrule

\tiny{GT} &
\imgclip{0}{img/synth/gt/48258-1-4-3__0.png} & 
\imgclip{0}{img/synth/gt/48258-1-4-3__1.png} &
\imgclip{0}{img/synth/gt/48258-1-4-3__2.png} &
\imgclip{0}{img/synth/gt/48258-1-4-3__3.png} &
\imgclip{0}{img/synth/gt/101943-0-4-2+bg1__0.png} &
\imgclip{0}{img/synth/gt/103778-0-3+bg0__0.png} &
\imgclip{0}{img/synth/gt/48379-0-2-4+bg3__0.png} &
\imgclip{0}{img/synth/gt/48379-0-2-4+bg3__1.png}\\

\tiny{\opdnetocs} &
\imgclip{0}{img/synth/ocv0_rgbd/48258-1-4-3__2.png} & 
\imgclip{0}{img/synth/ocv0_rgbd/48258-1-4-3__0.png} &
\imgclip{0}{img/synth/ocv0_rgbd/48258-1-4-3__1.png} &
\imgclip{0}{img/synth/ocv0_rgbd/48258-1-4-3__3.png} &
\imgclip{0}{img/synth/ocv0_rgbd/101943-0-4-2+bg1__0.png} &
\imgclip{0}{img/synth/ocv0_rgbd/103778-0-3+bg0__0.png} &
\imgclip{0}{img/synth/ocv0_rgbd/48379-0-2-4+bg3__0.png} &
\imgclip{0}{img/synth/ocv0_rgbd/48379-0-2-4+bg3__1.png} \\

\tiny{\ancsh} & Miss & Miss & Miss & Miss & 
\imgclip{0}{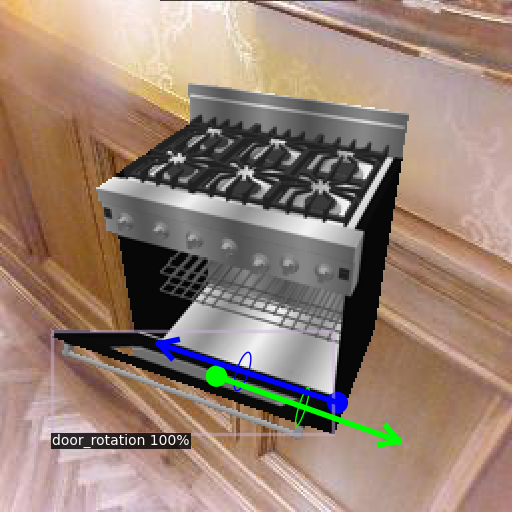} & 
\imgclip{0}{img/synth/ocv0_rgbd/103778-0-3+bg0__0.png} & Miss & Miss
\\

\tiny{\pnetopd} &
Miss &
\imgclip{0}{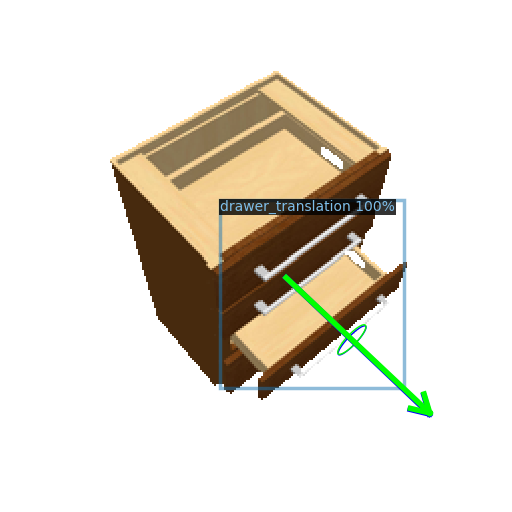} &
\imgclip{0}{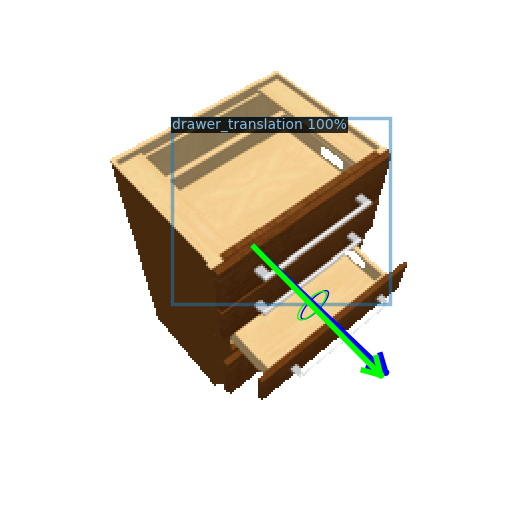} &
Miss &
\imgclip{0}{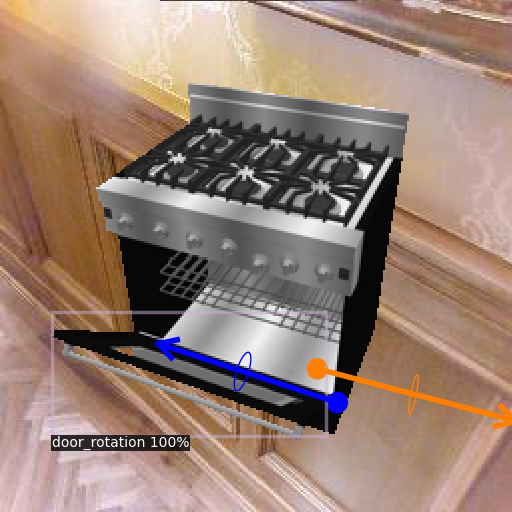} &
\imgclip{0}{img/synth/ocv0_rgbd/103778-0-3+bg0__0.png} & 
\imgclip{0}{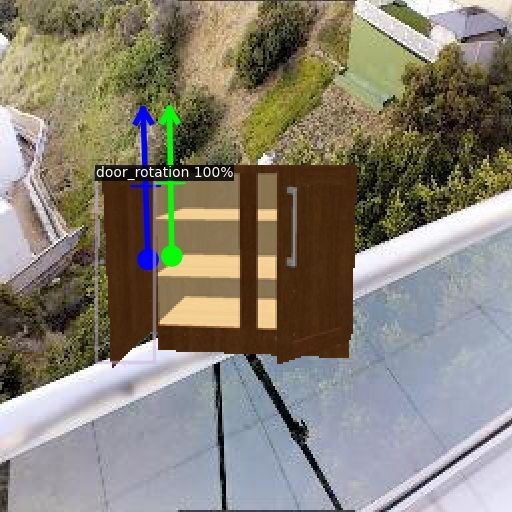} &
\imgclip{0}{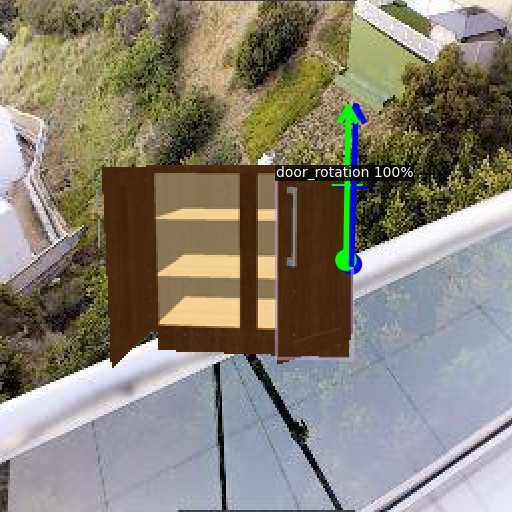} 
\\

\toprule
\multicolumn{9}{c}{\ourdatareal}\\
\midrule

\tiny{GT} &
\imgclip{0}{img/real/gt/425-660__0.png} &
\imgclip{0}{img/real/gt/425-660__1.png} &
\imgclip{0}{img/real/gt/425-660__2.png} &
\imgclip{0}{img/real/gt/657-300__1.png} &
\imgclip{0}{img/real/gt/232-60__0.png} &
\imgclip{0}{img/real/gt/232-60__1.png} &
\imgclip{0}{img/real/gt/232-60__2.png} &
\imgclip{0}{img/real/gt/191-3360__1.png} \\

\tiny{\opdnetocs} &
\imgclip{0}{img/real/ocv0_rgbd/425-660__0.png} &
\imgclip{0}{img/real/ocv0_rgbd/425-660__1.png} &
\imgclip{0}{img/real/ocv0_rgbd/425-660__2.png} &
\imgclip{0}{img/real/ocv0_rgbd/657-300__1.png} &
\imgclip{0}{img/real/ocv0_rgbd/232-60__0.png} &
\imgclip{0}{img/real/ocv0_rgbd/232-60__1.png} &
\imgclip{0}{img/real/ocv0_rgbd/232-60__2.png} &
\imgclip{0}{img/real/ocv0_rgbd/191-3360__0.png} \\

\tiny{\pnetopd} & Miss &
\imgclip{0}{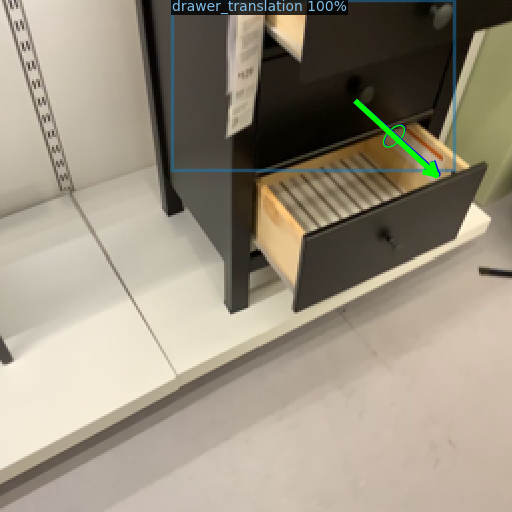} &
Miss &
\imgclip{0}{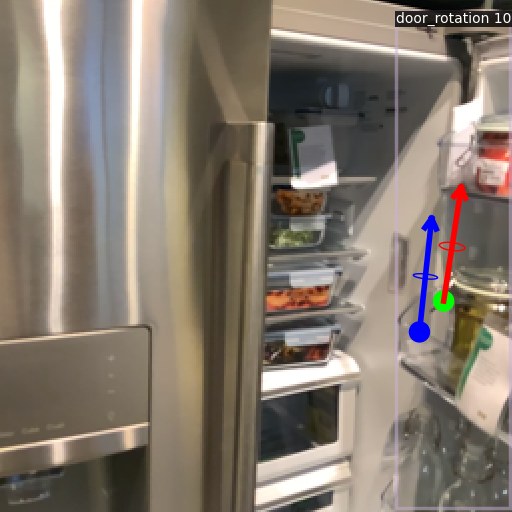} &
Miss &
\imgclip{0}{img/real/ocv0_rgbd/232-60__1.png} &
Miss &
\imgclip{0}{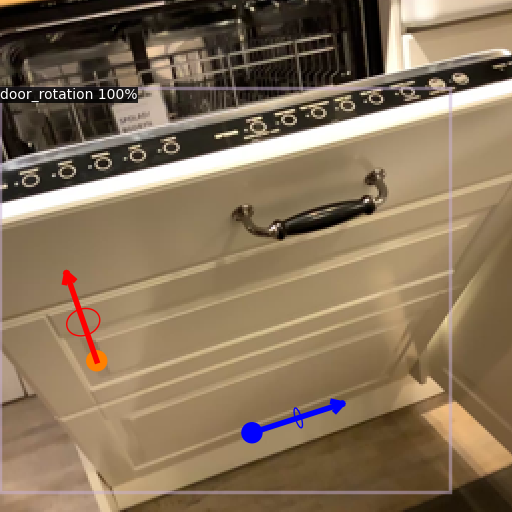}\\

\tiny{GT} &
\imgclip{0}{img/real/gt/477-1380__0.png} &
\imgclip{0}{img/real/gt/477-1380__1.png} &
\imgclip{0}{img/real/gt/450-540__0.png} &
\imgclip{0}{img/real/gt/450-540__1.png} &
\imgclip{0}{img/real/gt/218-180__0.png} &
\imgclip{0}{img/real/gt/218-180__1.png} &
\imgclip{0}{img/real/gt/218-180__2.png} &
\imgclip{0}{img/real/gt/218-180__3.png} \\

\tiny{\opdnetocs} &
\imgclip{0}{img/real/ocv0_rgbd/477-1380__1.png} &
\imgclip{0}{img/real/ocv0_rgbd/477-1380__0.png} &
\imgclip{0}{img/real/ocv0_rgbd/450-540__0.png} &
\imgclip{0}{img/real/ocv0_rgbd/450-540__1.png} &
\imgclip{0}{img/real/ocv0_rgbd/218-180__1.png} &
\imgclip{0}{img/real/ocv0_rgbd/218-180__0.png} &
\imgclip{0}{img/real/ocv0_rgbd/218-180__3.png} &
\imgclip{0}{img/real/ocv0_rgbd/218-180__2.png} \\

\tiny{\pnetopd} & Miss & Miss &
\imgclip{0}{img/real/gt/450-540__0.png} &
Miss &
Miss & Miss & Miss & Miss\\

\bottomrule
\end{tabularx}
\caption{Qualitative results comparing our approach against the \ancsh and \pnetopd baselines. Structure of the figure is the same as in the main paper. Both \ancsh and \pnetopd fail to detect many of the openable parts, in particular for the more challenging \ourdatareal dataset. In contrast, our \opdnetocs approach detects more parts and provides more accurate motion parameter estimates.}
\label{fig:results-prior}
\end{figure*}

\Cref{fig:results-prior} shows a qualitative comparison between \ancsh, \pnetopd and \opdnetoc.
We see that our \opdnetoc approach detects openable parts much more reliably, and overall provides more accurate motion parameter estimates for the detected parts.

\section{Dataset details}
\label{sec:supp:data}

We provide additional statistics on the part and structure variation (\Cref{sec:supp:data:statistics}) found in \ourdatacad and \ourdatareal, and details on how we rendered or selected images for the two datasets(\Cref{sec:supp:data:cad,sec:supp:data:real}).

\subsection{Dataset statistics}
\label{sec:supp:data:statistics}

\begin{table*}[t]
\caption{Openable part labels from \pmdataset~\cite{xiang2020sapien} for each object category we use in our experiments.}
\centering
\resizebox{\linewidth}{!}{
\begin{tabular}{@{} ll @{}}
\toprule
Category & Part labels\\
\midrule
Storage & cabinet\_door, drawer, drawer\_box, cabinet\_door\_surface, handle, glass, other\_leaf, door\\
Table & drawer, drawer\_box, handle, cabinet\_door, cabinet\_door\_surface, shelf, keyboard\_tray\_surface\\
Fridge & door, door\_frame, display\_panel, control\_panel, glass\\
Microwave & door\\
Washer & door\\
Dishwasher & door, door\_frame, display\_panel\\
Bin & cover, lid, frame\_vertical\_bar, opener, cover\_lid, other\_leaf, drawer\\
Oven & door, door\_frame\\
Safe & door\\
Box & rotation\_lid, lid\_surface, countertop, drawer\\
Suitcase & lid\\
\bottomrule
\\
\end{tabular}
}
\label{tab:dataset-initial-partlabels}
\end{table*}

\subsubsection{Part and motion statistics}
In \Cref{tab:dataset-part-stats} we report the total numbers of different part types, as well as number of images of different part types and number of motion types observed across all images.

\begin{figure*}
  \centering
  \includegraphics[width=\textwidth]{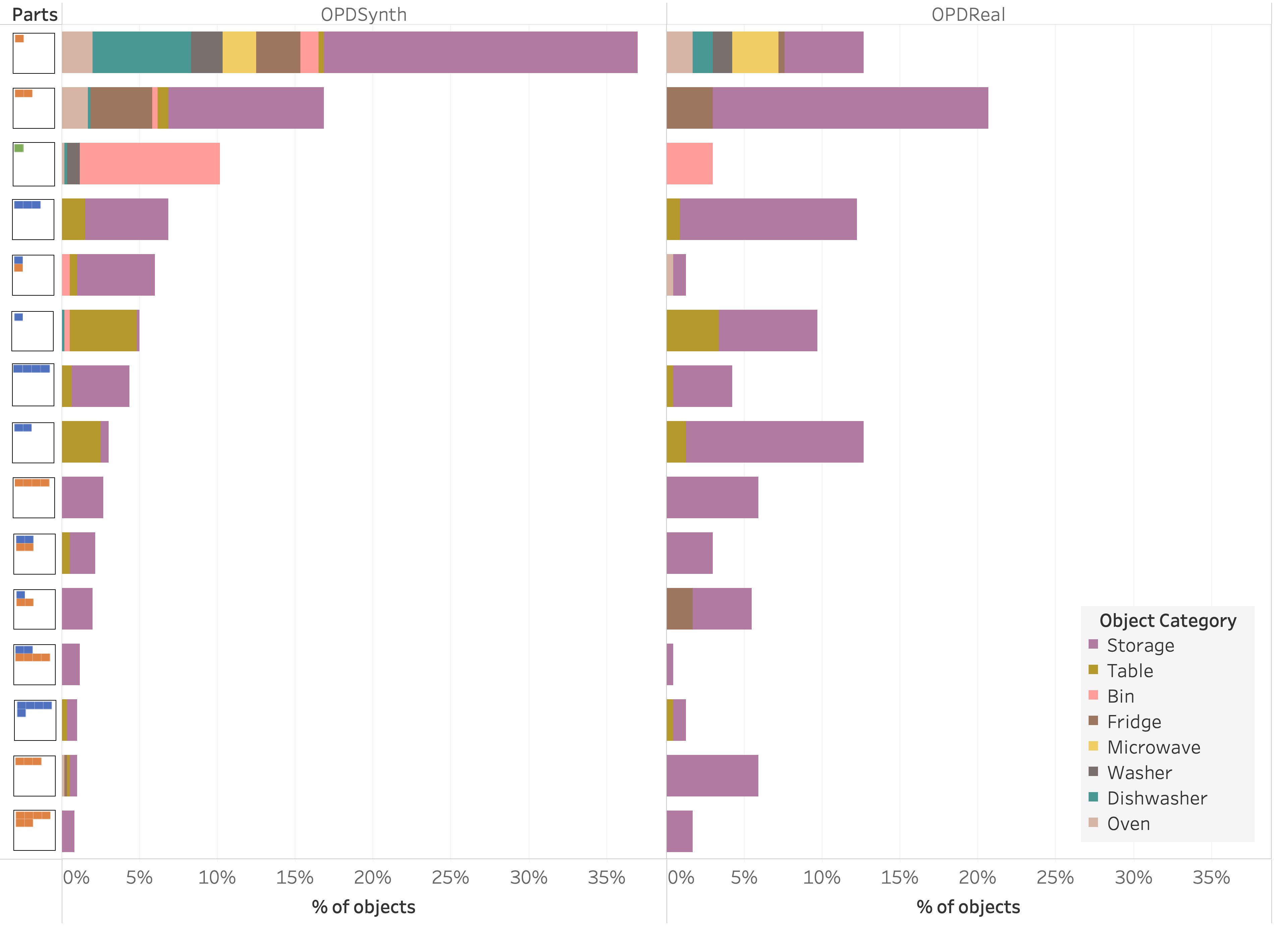}
  \caption{Part composition distribution for our \ourdatacad (left) and \ourdatareal (right) datasets. Each row represents a particular part composition with the icon at the left indicating the number of doors (orange), drawers (blue), and lids (green). The plot bar colors indicate the distribution over object categories. The top 15 part compositions are plotted sorted by number of objects with that part composition in the \ourdatacad dataset. We see that our datasets exhibit a diverse set of part compositions with varying numbers of parts of each type. These compositions are also distributed across several object categories.}
  \label{fig:structure}
\end{figure*}

\subsubsection{Openable part structure variation statistics}

We note that it is possible for objects belonging to the same object category to have variation in the structure (number of {\door}s, {\drawer}s, and {\lidd}s).  \Cref{fig:structure} shows the statistics of the structure variation of objects in our datasets.

\subsection{\ourdatacad details}
\label{sec:supp:data:cad}

\subsubsection{Consistent part labeling}
\label{sec:supp:part-labeling}

Our \ourdatacad dataset is based on synthetic objects from the \pmdataset dataset.
As the initial part labels for these objects may be inconsistent, we developed a two-pass approach to identify and label all openable parts.
In the first pass, we identify all part labels that may correspond to openable parts.
For each object category we identify the set of openable part labels from the set of all part labels.
For instance, for the object category of `box', we include `rotation lid', `lid\_surface' as openable parts, but not `base\_body' or `handle'. After collecting all part labels for each model category, we select an example model for each part label in that model category. Then through verifying the corresponding example model, we determine if we want to include this part label or not. From the accepted part labels (\Cref{tab:dataset-initial-partlabels}), we see that the semantic meaning of these part labels are all relevant to drawers, doors or lids. After the first pass, we get 740 models and 1441 parts over 11 categories.
The second pass is to verify all the parts selected from the first pass manually. We designed a user interface 
to show the part mobility and help annotators judge if it is a valid part which can be opened and closed.
We also relabel the parts with consistent labels from three main categories (\drawer, \door and \lidd). The annotation process took approximately 30 minutes to obtain 1343 valid parts and reassign them to the consistent set of openable part labels.

\begin{table*}[t]
\caption{Statistics of the distribution over part types and motion types in our datasets. The first three columns report the number of part instances (i.e. counting distinct openable parts across all objects). The second set of three columns reports the number of images with parts of that type across all objects. The last two columns report the number of images with a part exhibiting the specific motion type, across all parts and objects.}
\centering
\resizebox{\linewidth}{!}{
\begin{tabular}{@{} ll rrr rrr rr @{}}
\toprule
 & & \multicolumn{3}{c}{\# parts} & \multicolumn{3}{c}{\# part images} & \multicolumn{2}{c}{\# motion images}\\
 & & \drawer & \door & \lidd & \drawer & \door & \lidd & translation & rotation \\
 \cmidrule(l{0pt}r{2pt}){3-5} \cmidrule(l{2pt}r{2pt}){6-8} \cmidrule(l{2pt}r{0pt}){9-10}
\multirow{3}{*}{\ourdatacad}
 & train & 363 & 508 & 89 & 16882 & 141180 & 16880 & 171705 & 155175  \\
 & val & 79 & 94 & 20 & 28265 & 22635 & 4605 & 29200 & 26305   \\
 & test & 75 & 97 & 18 & 28425 & 25640 & 4825 & 28665 & 30225\\
\cmidrule(l{0pt}r{0pt}){2-10}
\multirow{3}{*}{\ourdatareal} 
 & train & 304 & 268 & 3 & 27598 & 17695 & 212 & 27540 & 17965  \\
 & val & 78 & 79 & 2 & 7324 & 5311 & 152 & 7520 & 5267   \\
 & test & 74 & 65 & 2 & 7002 & 4685 & 110 & 7002 & 4795 \\
\bottomrule
\end{tabular}
}
\label{tab:dataset-part-stats}
\end{table*}

\subsubsection{View selection}
\label{sec:supp:view-selection}

We render RGBD images of the object using a perspective camera, with a $50^\circ$ vertical field-of-view at $256 \times 256$ resolution.
For the RGB color we use Phong shading and a hemisphere light.
We vary the position and distance of the camera so that we get mostly views above and in front of the object.
To select specific views we sample the elevation $\theta$, azimuth $\phi$ and distance $d$ independently.
For each, we use a Bates distribution $B_k(a, b) = a + \frac{b-a}{k} \sum_{i=1}^k u_i$, that is the sum of $k$ standard uniform random variables $u_i$ scaled to the range $(a,b)$, with $a,b$ set as described below.
We first sample a categorical variable $v$ to determine if we want a camera viewpoint: 1) in a mostly above and frontal view with probability $0.6$; 2) in a wider vertical distribution for the camera allowing for the elevation $\theta$ to range from slightly below the object to above with probability $0.2$; or 3) in a wider horizontal distribution for the camera allowing for a larger range for the azimuth $\phi$ and distance $d$.
Note that the camera directly in front of the object is at $\phi=0$. 
The three cases are summarized below:
\begin{enumerate}\denselist
\item Above, frontal view:
\\$P(v=1)=0.6$, $\theta \sim B_2(30, 70), \phi \sim B_2(-60, 60), d \sim B_2(1.8, 2.8)$
\item Slightly below above:
\\$P(v=2)=0.2$, $\theta \sim B_3(-35, 35), \phi \sim B_2(-60, 60), d \sim B_2(1.8, 2.8)$
\item Slightly below above, wider azimuth/distance distribution:
\\$P(v=3)=0.2$ $\theta \sim B_3(-35, 35), \phi \sim B_3(-90, 90), d \sim B_2(1.6, 3.1)$
\end{enumerate}

\subsubsection{Image rendering}
\label{sec:supp:rendering}
We render single-view RGB and depth images from \ourdatacad.
For each object, we render a total of $5 + 20 \cdot \text{num\_parts}$ views each with different motion states for each part.
In one motion state all parts are at the min value of their motion range.
Then, we pick four random states for each moving part except the min value (one of which must be the max value of the range), while other parts stay at the min value of their motion range.
We augment the images using RGBD backgrounds from the Matterport3D~\cite{chang2017matterport3d} dataset by randomly selecting from the `straight ahead' and `downward tilt' camera views.
Each image has four random backgrounds resulting in a total of $25 + 100 \cdot \text{num\_parts}$ images.

\subsection{\ourdatareal details}
\label{sec:supp:data:real}

\subsubsection{Data capture and reconstruction}
We used iPad Pro 2021 devices to capture RGB-D video scans of articulated objects in indoor environments.
We focused on object categories that overlap with \ourdatacad and have openable parts.
Each scan focused on a single object instead of capturing the entire environment.
We take multiple scans ($\sim3$) of each object.
In total, three student volunteers collected 863 scans covering 294 different objects across 8 object categories.
From these, we obtained 763 polygonal meshes for 284 different objects (some scans failed to produce high quality reconstructions).
We obtained polygonal mesh reconstructions from these scans using the Open3D~\cite{zhou2018open3d} implementation of RGB-D integration and \citet{waechter2014texturing}'s implementation of texturing.

\subsubsection{Annotation}
We adapted the 3D annotation tool from ScanNet~\cite{dai2017scannet} to work with textured meshes and used it to annotate object parts (`door', `drawer', `lid', or `base').
For the object articulation, we use the annotation interface from \citet{xu2020motion}.
Each articulatable part is annotated with the motion type (`rotation' or `translation'), motion axis and rotation origin, as well as motion ranges.
We developed another interface to indicate the semantic orientation (front) of the oriented bounding box for the object.
The annotation was done by student volunteers.
The semantic part annotation took the longest with an average of 7.18 minutes per object, and 5483 minutes in total (across 5 volunteers).
Articulated part annotation took an average of 3.24 minutes per object and 2473 minutes in total.
The annotation of semantic OBBs was much faster with an average of $\sim20$ seconds taken per scan.
Articulation and semantic OBB annotation was done by one of the authors.

\subsubsection{Frame selection}
For \ourdatareal, we selected frames from each scan to form our image dataset.
We rescale both the depth and color frames to a 256 x 256 resolution using a center-crop strategy.
When selecting frames, we sample one frame every second ensuring that at least 1\% of pixels belong to an openable part and at least 20\% of parts are visible.

\end{document}